%% file: collas2023_conference.tex
\documentclass{article} 
\usepackage{collas2023_conference,times}
\usepackage{caption}
\usepackage{subcaption}
\usepackage{misc_symbols}
\usepackage{wrapfig}
\usepackage{authblk}

\usepackage[export]{adjustbox}
\usepackage{easyReview}

\input{math_commands.tex}

\usepackage{hyperref}
\hypersetup{
    colorlinks=true,
    linkcolor=red,
    filecolor=magenta,
    urlcolor=blue,
    citecolor=purple,
    pdftitle={Overleaf Example},
    pdfpagemode=FullScreen,
    }

\title{What Happens During Finetuning of Vision Transformers: An Invariance Based Investigation}

\author[1]{Gabriele Merlin}
\author[1,2]{Vedant Nanda}
\author[1]{Ruchit Rawal}
\author[1]{Mariya Toneva}
\affil[ ]{\texttt{\{gmerlin,vnanda,rrawal,mtoneva\}@mpi-sws.org}}
\affil[1]{\footnotesize MPI-SWS, Germany}
\affil[2]{\footnotesize University of Maryland, USA}

\collasfinalcopy 

\begin{document}

\maketitle

\begin{abstract}
The pretrain-finetune paradigm usually improves downstream performance over training a model from scratch on the same task, becoming commonplace across many areas of machine learning. While pretraining is empirically observed to be beneficial for a range of tasks, there is not a clear understanding yet of the reasons for this effect. In this work, we examine the relationship between pretrained vision transformers and the corresponding finetuned versions on several benchmark datasets and tasks. We present new metrics that specifically investigate the degree to which invariances learned by a pretrained model are retained or forgotten during finetuning. Using these metrics, we present a suite of empirical findings, including that pretraining induces transferable invariances in shallow layers and that invariances from deeper pretrained layers are compressed towards shallower layers during finetuning. Together, these findings contribute to understanding some of the reasons for the successes of pretrained models and the changes that a pretrained model undergoes when finetuned on a downstream task.

\end{abstract}

\section{Introduction}
\label{sec:introduction}
In recent years much progress in deep learning has been driven by the reuse of models that were pretrained on large amounts of data. This is usually achieved by finetuning their parameters using a smaller amount of data from a target downstream task.  This pretrain-finetune paradigm usually improves downstream performance over training a model from scratch on the same task, and has become commonplace across many areas of machine learning, including natural language processing \citep{howard-ruder-2018-universal} and computer vision \citep{girshick2014rich}. 

While pretraining is empirically observed to be beneficial for a range of tasks, there is not a clear understanding yet of the reasons for this effect. 
Previous work has empirically examined various conditions for pretraining and found that for a given budget of pre-training images, training with fewer classes, but more images per class performs better \citep{huh2016makes}. 
Pretraining has also been posited to elicit an accelerated convergence during finetuning \citep{kornblith2019better}, suggesting that during pretraining, models learn transferable representations, particularly when the finetuning task domain is similar to the pretraining task.

In this work, we further examine the relationship between pretrained vision transformers and the corresponding finetuned versions on several benchmark tasks and datasets. A key to our study is that we leverage STIR \citep{nanda2022measuring}, a recent approach that estimates how much of the invariances to specific perturbations learned by one source model are shared with a second target model. We adopt this approach because of the observation that learning to be invariant to some perturbations has been shown to improve generalization ability in individual models \citep{devries2017improved, zhang2017mixup,yun2019cutmix}, and the transfer learning ability to other models ~\citep {salman2020adversarially}. This suggests that learning invariant representations may enable generalization, and so it is possible that pretrained models are learning such invariant representations. STIR can help understand the extent to which invariance from the pretrained model is learned or forgotten by a finetuned model.

Using this approach, we define metrics that are useful for tracking the degree to which pretrained invariances are forgotten and new invariances are learned by finetuning a pretrained model. A well-known constraint that occurs during training is the stability-plasticity dilemma \citep{mermillod2013stability}, which refers to the trade-off between the ability of a neural network to retain information that has already been learned (stability) and the ability to learn new information (plasticity). Finding a balance between these two factors is thought to be crucial for the successful functioning of a neural network. Developing metrics that can capture the degree to which pretrained invariances are forgotten and new invariances are learned during finetuning allows us to characterize the trade-off between old and new invariances during the pretrain-finetune paradigm. 

Using these metrics, we present a suite of empirical findings, including that pretraining induces transferable invariances, especially in the shallow layers of the network (\ie\ closer to the inputs), and that invariances from deeper pretrained layers are \emph{compressed} towards shallower layers during finetuning. Together, these findings contribute to understanding some of the reasons for the successes of pretrained models and the changes that a pretrained model undergoes when finetuned on a downstream task.

\section{Related Works}
\label{sec:related_works}
\paragraph{Forgetting and learning.}
Forgetting and learning have been studied extensively in continual learning \citep{lesort2020continual,kirkpatrick2017overcoming,kemker2018measuring}. In this setting, a model is trained on a sequence of tasks and is required to maintain performance on previously learned tasks while learning new tasks. \citet{toneva2018empirical} study data instances that are learned and forgotten many times during the training process. 
These previous works primarily measure forgetting and learning from a behavioural perspective: they rely on task performance to quantify these measures and do not take into account invariances. In our work, we provide a fresh perspective on learning and forgetting through the lens of invariances. This unique lens allows us to propose two additional notions of compression and expansion, which provide a more complete picture of how representations change during finetuning and something that prior works have not considered. There are three notable works that go beyond task performance and are similar in spirit to ours:~\citet{ramasesh2020anatomy} use representation similarity and \citet{davari2022probing} use linear probes, both to understand catastrophic forgetting in a continual learning setup. More recently, \cite{ramasesh2022effect} investigated the role of pretraining scale in learning orthogonal class representations that lead to lower catastrophic forgetting during continual learning. We differ in two key aspects: we consider the setup of transfer learning which is more broadly applicable to a variety of machine learning tasks; and in addition to just using representations (via either representation similarity or linear probes), we also measure shared invariances that allow us to derive additional insights about changes that occur due to finetuning. 

\paragraph{Using Similarity to Study Representations.}
In recent years, numerous works have adopted representational similarity measures (RSMs) to inspect representations of neural networks that differ in architecture family \citep{raghu2021vision}, depth/width \citep{nguyen2020wide}, and localization of information \citep{wu-etal-2020-similarity}. Most similar to our work, \cite{phang-etal-2021-fine} utilize RSMs to investigate the changes in representations of a finetuned model with respect to the pretrained model. While conventional RSMs are useful in indicating 'representational-divergence' (\ie\ how are the representations changing between two models for the same set of inputs), they are not equipped to quantify the degree to which two models share invariances. Thus, we build upon an approach proposed by \citet{nanda2022measuring}, which reveals the extent to which the learned representations remain invariant to the same perturbation from one model to another. We analyze the pretraining-finetuning paradigm of ViT models that was not evaluated in \citet{nanda2022measuring}. Moreover, we propose novel metrics that probe the extent and nature of \emph{forgetting} and \emph{learning} of shared-invariances due to finetuning.

\section{Methods}
\label{sec:methods}
We build upon a recently proposed general approach to estimate the shared invariances between two models, which we describe in Section \ref{subsec:methods_stir}. In Sections \ref{subsec:methods_forgetting_and_learning} and \ref{subsec:methods_compression_and_expansion}, we define metrics that are useful to characterize the forgetting of pretrained invariances and learning of new invariances by the finetuned model. In Section \ref{subsec:methods_models}, we describe the experimental setup used to validate these metrics and to use them to reveal new insights about how models change during finetuning. Invariance is used in many contexts in the broad ML literature, however, here we adopt the terminology of~\cite{nanda2022measuring} and use the word invariance to broadly mean \textit{transformations of inputs that do not significantly change the representation of a model at a particular layer}.

\subsection{Similarity Through Inverted Representations (STIR)}
\label{subsec:methods_stir}
Similarity Through Inverted Representations (STIR) \citep{nanda2022measuring} is a measure of shared invariances between two representations.
To measure the degree to which the $i^{th}$ layer of model $m_2$ shares invariances with the $j^{th}$ layer of model $m_1$, \citet{nanda2022measuring} define STIR as:

 \begin{equation}\label{eq:stir}
    \texttt{STIR}(m^{[i]}_{2} | m^{[j]}_{1}, X, S_{r}) = \frac{1}{k}\sum_{X'}S_{r}(m^{[i]}_{2}(X), m^{[i]}_{2}(X')).
\end{equation}

In equation \ref{eq:stir}, $X$ is a set of samples and $X'$ is a set of generated samples such that $m^{[j]}_{1}(X) \approx m^{[j]}_{1}(X')$ - \ie\ a set of perturbed samples for which the $j^{th}$ layer of $m_1$ is representationally invariant. $S_{r}$ is a similarity metric, which is often taken to correspond to CKA \citep{kornblith2019similarity}. Thus, using STIR one can compute the degree of shared-variances between the representations learned by any two layers ($i$ \& $j$) both across two models ($ \texttt{STIR}(m^{[j]}_{2} | m^{[i]}_{1}, X, S_{r})$) or within an individual model ($ \texttt{STIR}(m^{[j]}_{1} | m^{[i]}_{1}, X, S_{r})$). Note that STIR of a layer with itself within an individual model (\ie\ $ \texttt{STIR}(m^{[i]}_{1} | m^{[i]}_{1}, X, S_{r})$) is 1. The sampling of $X$ is repeated $k$ times. Unlike the standard CKA that captures changes in the representation between two models, STIR is able to capture the robustness of a target model to perturbations on which a reference model is representationally invariant. Note that STIR is directional: $\texttt{STIR}(m^{[i]}_{2} | m^{[j]}_{1}, X, S_{r})\neq \texttt{STIR}(m^{[j]}_{1} | m^{[i]}_{2}, X, S_{r})$.
For our purposes, STIR is useful for disentangling the \texttt{learning} and \texttt{forgetting} of invariances in a finetuned model, which cannot be easily disentangled using standard CKA. It is also useful to compare layers with different sizes, since it is based on CKA which by design has a normalization factor that ensures invariance to isotropic scaling \citep{kornblith2019similarity,nanda2022measuring}%

STIR uses representations similarity (CKA by \cite{kornblith2019similarity}) under the hood – which by design has a normalization factor that ensures invariance to isotropic scaling. This makes both representation similarity (CKA) and STIR suitable for comparison across layers of different sizes. Comparison of CKA and STIR across layers was also done by both \cite{kornblith2019similarity} and \cite{nanda2022measuring}. Thus we believe there’s already proper normalization in both STIR and CKA which make them suitable for cross-layer comparisons.

\subsection{Forgetting and Learning}
\label{subsec:methods_forgetting_and_learning}

To measure the extent to which the model learns and forgets invariances during finetuning, we propose two metrics: \texttt{learning} and \texttt{forgetting}. Unlike previous work described in Section \ref{sec:related_works}, our aim is to characterize them from a representational robustness perspective using the STIR measure. 

We define the $\texttt{forgetting}(ft^{[i]}|pt^{[j]})$, where $ft^{[i]}$ is a layer of the finetuned neural network as:
\begin{equation}
\label{eq:forgetting}
    \texttt{forgetting}(ft^{[i]}|pt^{[j]})=\texttt{STIR}(pt^{[i]}|pt^{[j]})-\texttt{STIR}(ft^{[i]}|pt^{[j]}).
\end{equation}

In the second term of Equation \ref{eq:forgetting}, we measure the shared invariances between the layer $i$ of the finetuned model ($ft$) and the layer $j$ of the pretrained model ($pt$). Thus, intuitively $\texttt{forgetting}(ft^{[i]}|pt^{[j]})$ measures the decrease in shared-invariances between $i^{th}$ and $j^{th}$ layers after finetuning (\ie\ $\texttt{STIR}(ft^{[i]}|pt^{[j]})$) relative to after pretraining (\ie\  $\texttt{STIR}(pt^{[i]}|pt^{[j]})$).
We are interested in measuring the evolution of the same layer during finetuning. Therefore the \texttt{forgetting} of a finetuned layer ($i=j$) is measured by $\texttt{forgetting}(ft^{[i]}|pt^{[i]})=\texttt{STIR}(pt^{[i]}|pt^{[i]})-\texttt{STIR}(ft^{[i]}|pt^{[i]})$. If the first term, which is always 1, is greater than the second we can say that layer $i$ is forgetting.

Similarly we define $\texttt{learning}(ft^{[i]}|pt^{[j]})$ as:
\begin{equation}
\label{eq:learning}
    \texttt{learning}(ft^{[i]}|pt^{[j]})=\texttt{STIR}(ft^{[i]}|ft^{[j]})-\texttt{STIR}(pt^{[i]}|ft^{[j]}).
\end{equation}

In the second term of Equation \ref{eq:learning}, we measure the shared invariances between the layer $j$ of the finetuned model ($ft$) and layer $i$ of the pretrained model ($pt$). Intuitively, if the invariances defined w.r.t a finetuned model are shared by a pretrained model (\ie\ $\texttt{STIR}(pt^{[i]}|ft^{[j]})$), the degree of new invariances learned during finetuning is low.
 The \texttt{learning} of a particular finetuned layer ($i=j$) is measured by $\texttt{learning}(ft^{[i]}|pt^{[i]})=\texttt{STIR}(ft^{[i]}|ft^{[i]})-\texttt{STIR}(pt^{[i]}|ft^{[i]})$. If the first term, which is always 1, is greater than the second we can say that layer $i$ is learning. These metrics can measure the relative levels of learning and forgetting between two models. As a result, they are useful for comparing the effect of different design choices  involved in finetuning: pretrained models, finetuning tasks, or datasets.

\citet{ramasesh2020anatomy} analyzed the phenomenon of catastrophic forgetting using representation similarity to identify the layers most responsible for forgetting.
To compare our STIR-based metrics with CKA, we propose a metric similar to \citet{ramasesh2020anatomy}. We define the \texttt{cka\:divergence} as:
\begin{equation}
\label{eq:cka_divergence}
    \texttt{cka\:divergence}(ft^{[i]},pt^{[j]})=\texttt{CKA}(pt^{[i]},pt^{[j]})-\texttt{CKA}(ft^{[i]},pt^{[j]}).
\end{equation}

Since we are interested in measuring the evolution of a specific layer ($i=j$), we can use $\texttt{cka\:divergence}(ft^{[i]},pt^{[i]})$. Therefore, the first term of this equation is always 1. As a result, we are measuring how much the representations of layer $i$ of the finetuned model ($ft$) vary from the pretrained model ($pt$). As result, when $\texttt{CKA}(ft^{[i]},pt^{[i]})$ is low the \texttt{cka\:divergence} is high. Since  \texttt{CKA} is not a directional metric, we cannot distinguish between learning and forgetting.
Our metric is qualitatively the opposite of the one used by \citet{ramasesh2020anatomy}.

\subsection{Compression and Expansion}
\label{subsec:methods_compression_and_expansion}

During finetuning, a model may not only forget or learn invariances, but the invariances from a specific layer in the pretrained model may also migrate to a different layer in a finetuned model. To measure the migration of invariances during finetuning, we propose \texttt{Invariance\:Flow}. For two layers $i$ and $j$ we define the \texttt{Invariance\:Flow}$(ft^{[i]},pt^{[j]})$ as:
\begin{equation}
\label{eq:invariance_flow}
    \texttt{Invariance\:Flow}(ft^{[j]},pt^{[i]})=\texttt{STIR}(ft^{[j]}|pt^{[i]})-\texttt{STIR}(ft^{[i]}|pt^{[i]})
\end{equation}

Equation \ref{eq:invariance_flow} contrasts the ability of layer $j$ of the finetuned model to share invariances with layer $i$ of the pretrained model, with the same ability but of layer $i$. The result of this metric is the \texttt{Invariance\:Flow\:Matrix}, which describes the flow of invariances from a pretrained layer to another finetuned layer. As a result, if $\texttt{Invariance\:Flow\:Matrix}(i,j)>0$ and $j<i$ we can say that the invariances of the pretrained model are \textit{compressed} to earlier layers, otherwise if $j>i$ they are \textit{expanded} to deeper layers. If $i=j$ the \texttt{Invariance\:Flow} is 0 since the two terms of the equation would be equal.

Our proposed metrics can be generalized to other settings \ie\ between any pair of reference and target models. We explicitly use the notation $ft$ and $pt$ for clarity. 

\subsection{Models and Tasks}
\label{subsec:methods_models}
\paragraph{Models.} For our experiments, we  use the Vision Transformer (ViT) model, proposed by \citet{dosovitskiy2020image}. ViT is a variant of the Transformer architecture \citep{vaswani2017attention} for images. The architecture of Vision Transformer is similar to the standard Transformer, consisting of a stack of multi-head self-attention layers followed by fully connected layers. The key difference is the use of convolutional layers for image feature extraction rather than the embedding layer used in the original Transformer. We use a ViT model with 12 layers that use a patch size of 32x32 and 224x224 as image resolution. ViT models have proven to be effective end efficient in many tasks, achieving good performance in a shorter training time with respect to convolutional models. We use the implementation of ViT provided by \citet{rw2019timm}.
For our experiment, ViT models are pretrained on ImageNet \citep{deng2009imagenet} classification or pretrained on CIFAR100 \citep{cifar} classification (see details in Appendix \ref{app:training_details}).

\paragraph{Finetuning Tasks and Datasets.} To understand how the difference between a pretraining and a finetuning task affects the shared invariances, we finetune the pretrained models on two different tasks: classification and reconstruction. Since all pretrained models that we use are trained using classification, the reconstruction task is helpful in analyzing changes in shared invariance that occur when the finetuning task is different from the pretraining task. For the reconstruction task, we reconstruct the original image from ViT representations using a convolutional decoder based on the Hugging Face implementation \citep{wolf2020transformers}. To train the model, we adopt the L1 loss as the objective function to ensure that the reconstructed image is as close as possible to the original image. For classification, we adopt the standard cross-entropy loss function. 
We performed experiments with 3 representative datasets from the Visual Task Adaptation Benchmark \citep{zhai2019large}, two naturalistic datasets - CIFAR100 \citep{cifar} and Oxford-IIT Pet \citep{pets_dataset} - and one specialized dataset - Eurosat \citep{helber2019eurosat}, a satellite imagery dataset. We further experimented with CIFAR10 \citep{cifar} to compare the results of CIFAR100 with a dataset of similar domain and to use a well-known benchmark widely used in the literature.
More details about training hyperparameters and methods are reported in Appendix \ref{app:training_details}.

\section{Results}
\label{sec:results}

Using the proposed metrics, we examine two main effects on the forgotten pretrained invariances and the newly learned invariances by the finetuned model: 1) the effect of the type of task a pretrained model is finetuned for (Section \ref{subsec:learning_and_forgetting_invariances}), and 2) the effect of the dataset on which the pretrained model is trained on (Section \ref{subsec:pretraining effect}). We further investigate whether invariances that appear to have been forgotten in a certain finetuned layer have actually migrated to a different layer in the model (either a shallower or a deeper layer) in Section \ref{subsec:copression_and_expansion_analysis}. We finally study the training dynamics of forgotten and learned invariances in Section \ref{subsec:training_dynamics}. All results are obtained by averaging over $3$ random seeds of STIR computation. For each STIR computation, we set the number of sampled images ($X$ in Equation \ref{eq:stir}) to 500, and we use $50$ optimization iterations to find a representationally invariant input for each sampled image.

\subsection{Effects of the finetuning task}
\label{subsec:learning_and_forgetting_invariances}

\begin{figure}[ht]
\centering
    \begin{subfigure}[b]{0.29\textwidth}

        \includegraphics[width=1\textwidth]{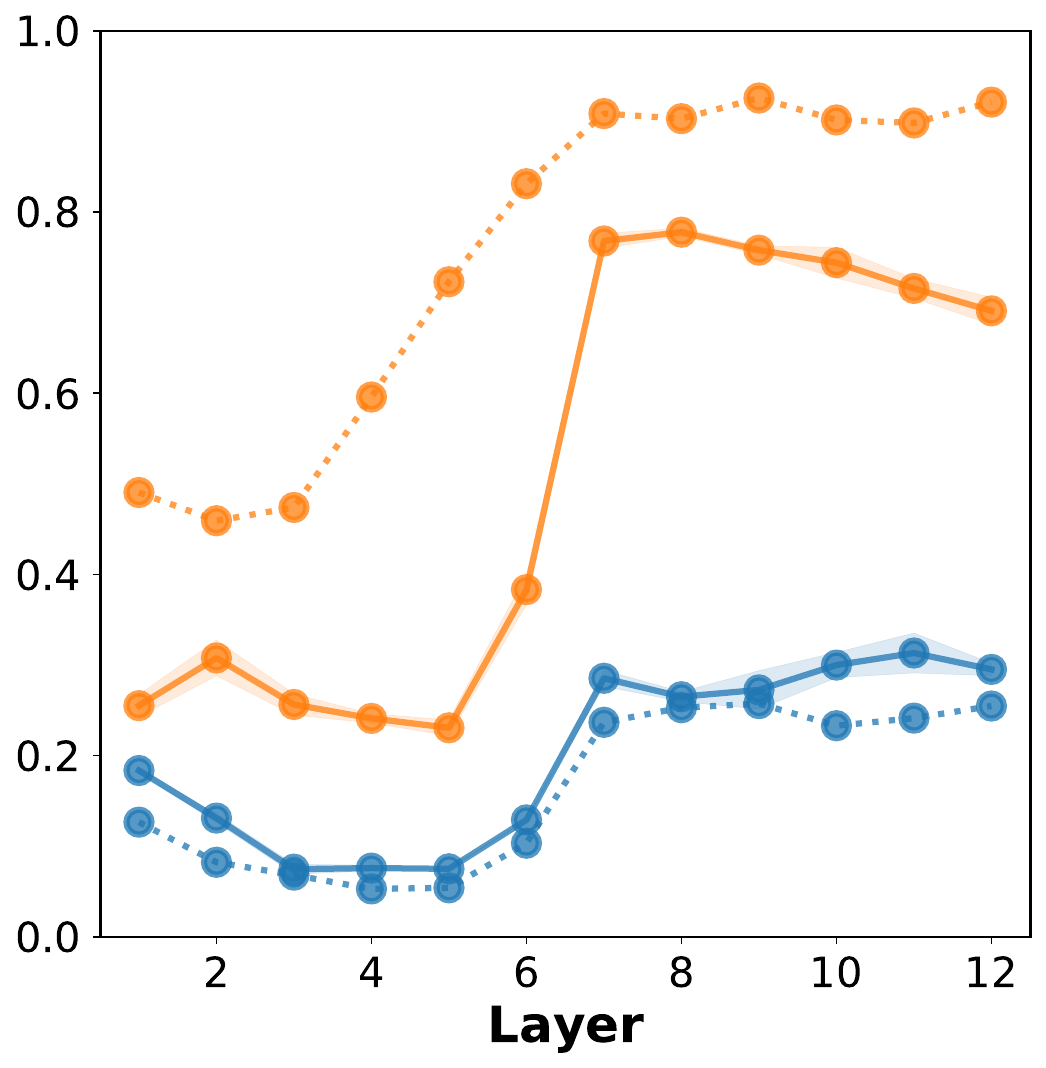}
        \caption{Learning and forgetting CIFAR10}
        \label{fig:learning_forgetting_cifar10}
    \end{subfigure}
    \captionsetup[subfigure]{labelformat=empty}
   \raisebox{3.5\height}{
       \begin{subfigure}[b]{0.18\textwidth}
            \includegraphics[width=1\textwidth]{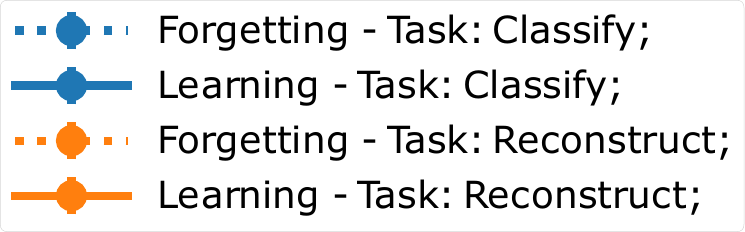}
        \end{subfigure}
    }
    \captionsetup[subfigure]{labelformat=parens}
    \begin{subfigure}[b]{0.29\textwidth}
        \includegraphics[width=1\textwidth]{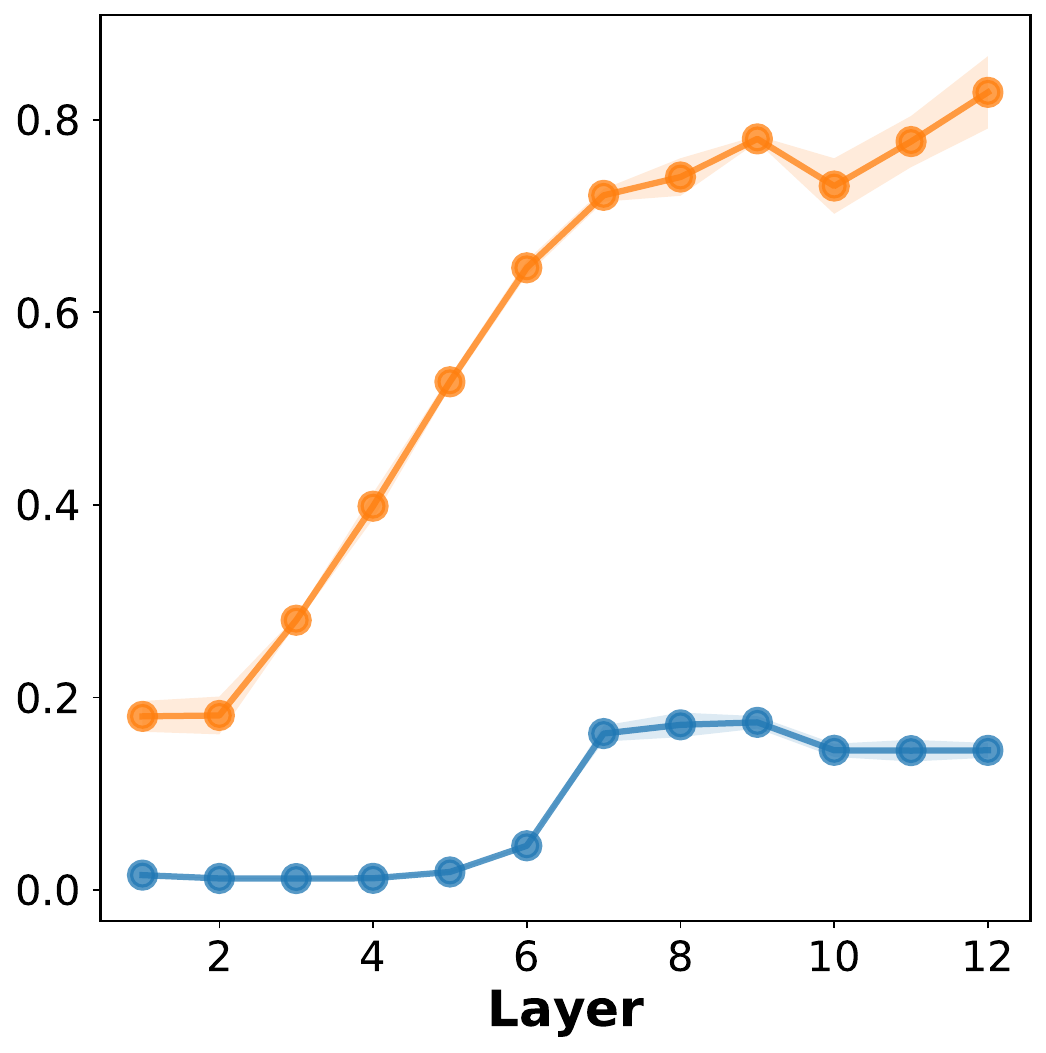}
        \caption{CKA divergence CIFAR10}
        \label{fig:cka_divergence_cifar10}
    \end{subfigure}
    \captionsetup[subfigure]{labelformat=empty}
    \raisebox{7\height}{
   \begin{subfigure}[b]{0.18\textwidth}
        \includegraphics[height=0.17\textwidth]{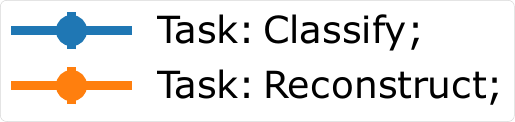}
    \end{subfigure}
    }
   \caption{\texttt{learning}, \texttt{forgetting} and \texttt{cka\:divergence} values for ViT model pretrained on ImageNet and finetuned on CIFAR10 on reconstruction and classification task. \texttt{learning}, \texttt{forgetting} measure two different quantities that may differ from each other. The model finetuned on the reconstruction task shows different levels and dynamics of \texttt{learning} and \texttt{forgetting} with respect to the model finetuned on the classification task. Changes in \texttt{cka\:divergence} vary from task to task. The model finetuned on the reconstruction task shows higher \texttt{cka\:divergence} in each layer. Later layers for both task have a higher \texttt{cka\:divergence} with respect to earlier layers.}
   \vspace{-4mm}
\end{figure}

To study the effect of different finetuning tasks on forgetting pretrained invariances and on learning new invariances during finetuning, we examine the \texttt{forgetting} and \texttt{learning} metrics across layers of a ViT model pretrained on ImageNet and finetuned on two tasks for the same CIFAR10 dataset: 1) classification and 2) reconstruction. We present the results in Figure \ref{fig:learning_forgetting_cifar10}. The results for the remaining datasets (CIFAR100, Oxford-IIT Pet, EuroSAT) are qualitatively similar and can be viewed in Appendix \ref{app:task_cifar100}. 

\paragraph{Learning and forgetting capture different phenomena.}
Intuitively, a model with a given capacity to store information may forget previously learned invariances in order to learn new ones. Therefore, the 
\texttt{forgetting} and \texttt{learning} metrics may be thought of as two consequences of the same phenomenon. However, the results in Figure \ref{fig:learning_forgetting_cifar10} show that the two corresponding metrics can lead to different results. In particular, the metrics behave differently when the task for pretraining (\ie\ classification) differs from the task for finetuning (\ie\ reconstruction). The clearest example is for layers $3-5$, where the \texttt{learning} during the reconstruction finetuning is constant while the \texttt{forgetting} is steadily increasing, and for layers $9-12$, where the \texttt{forgetting} during the reconstruction finetuning is constant while the \texttt{learning} is steadily decreasing. Therefore, the proposed \texttt{learning} and \texttt{forgetting} metrics capture some unique information and may reveal different insights.

\paragraph{Representational similarity does not imply shared invariance.} Analyzing Figure \ref{fig:learning_forgetting_cifar10}, we can see that the model finetuned for the classification task has moderate \texttt{forgetting} and \texttt{learning} values for the initials layers. The two metrics decrease up to layer 5 and then start to increase, reaching values that are slightly above the ones evaluated at the earlier layers. This means that the initial layers and the later layers are the ones where more \texttt{forgetting} and \texttt{learning} of invariances occurs during finetuning. In contrast, if we were to use a representational similarity measure to quantify the differences in representations between the finetuned model and the pretrained model, we observe a very different trend in the early layers for the classification task (see Figure \ref{fig:cka_divergence_cifar10}). The representational similarity in the early layers is very high, corresponding to almost $0$ values for the \texttt{cka\:divergence}. This difference between the \texttt{cka\:divergence} and the \texttt{forgetting} and \texttt{learning} metrics shows that 
representational similarity does not imply shared invariance. This result supports previous findings from \citet{nanda2022measuring}, that similarly show that high values of CKA (or low values of \texttt{cka\:divergence}) can correspond to various degrees of shared invariance. Therefore, the proposed metrics based on relative invariance may provide additional insight into how a model changes during finetuning.

\paragraph{Earlier layers \emph{do} change, even when the finetuning task matches the pretraining task.} Previous work based on representational similarity has reported that during finetuning, later layers adapt more to the finetuning task than earlier layers \citep{ramasesh2020anatomy}. Thus, we can expect not only that the representations change more in the later layers than in the earlier layers, but also that later layers learn new invariances and forget more pretrained invariances. Surprisingly, in Figure \ref{fig:learning_forgetting_cifar10}, we observe that the initial layers (0,1) also learn and forget invariances similarly to the later layers, thus adapting to the new finetuning task and data set. As the model undergoes changes in the input distribution during finetuning (from ImageNet to CIFAR10), we hypothesise that earlier layers need to adapt to the new input distribution. \cite{ramasesh2020anatomy} focused more on a continual learning setting and the result based on \texttt{forgetting} and \texttt{learning}  may differ. However we show in Figure \ref{fig:cka_divergence_cifar10} and also in a pretraing-finetuning setting CKA is not able to capture changes in the model.

\paragraph{Transferring to a new task can require learning new invariances.} 
Finetuning a model on a task that is different from the one on which it was trained may require learning new invariances, and to a larger degree than if the model was finetuned on the same pretraining task. We can examine this question by contrasting the \texttt{learning} values for the two tasks in Figure \ref{fig:learning_forgetting_cifar10}: classification and reconstruction.
We observe that new invariances are learned by both finetuned models. For the reconstruction task, the \texttt{learning} metric is higher in every layer compared to the classification task. This means that, as expected, finetuning a pretrained model on a new task may require learning new invariances.

\paragraph{Transferring to a lower-level task can require forgetting in earlier layers.}
The ability of deep neural networks to learn increasingly abstract concepts with increasing network depth has been widely studied in the literature \citep{zeiler2014visualizing}. Is this effect still observable when analysing learned and forgotten invariances during finetuning? As we can see in our experiments in Figure \ref{fig:learning_forgetting_cifar10}, a low-level task, such as reconstruction, starts the increasing trend of \texttt{forgetting} from layer 3. Instead, a more abstract task, such as classification, starts the increasing trend of \texttt{forgetting} from layer 5 onwards. This means that \texttt{forgetting} shows an earlier growth trend for the model finetuned to the reconstruction task, compared to the model finetuned on a more abstract task such as classification. We hypothesize that because the reconstruction task is a lower-level task, the model forgets earlier invariances in the hierarchy of layers.

\subsection{Effects of the pretraining dataset}
\label{subsec:pretraining effect}

\begin{figure}[t]
  \centering

    \begin{subfigure}[b]{0.35\textwidth}
        
        \includegraphics[width=1\textwidth]{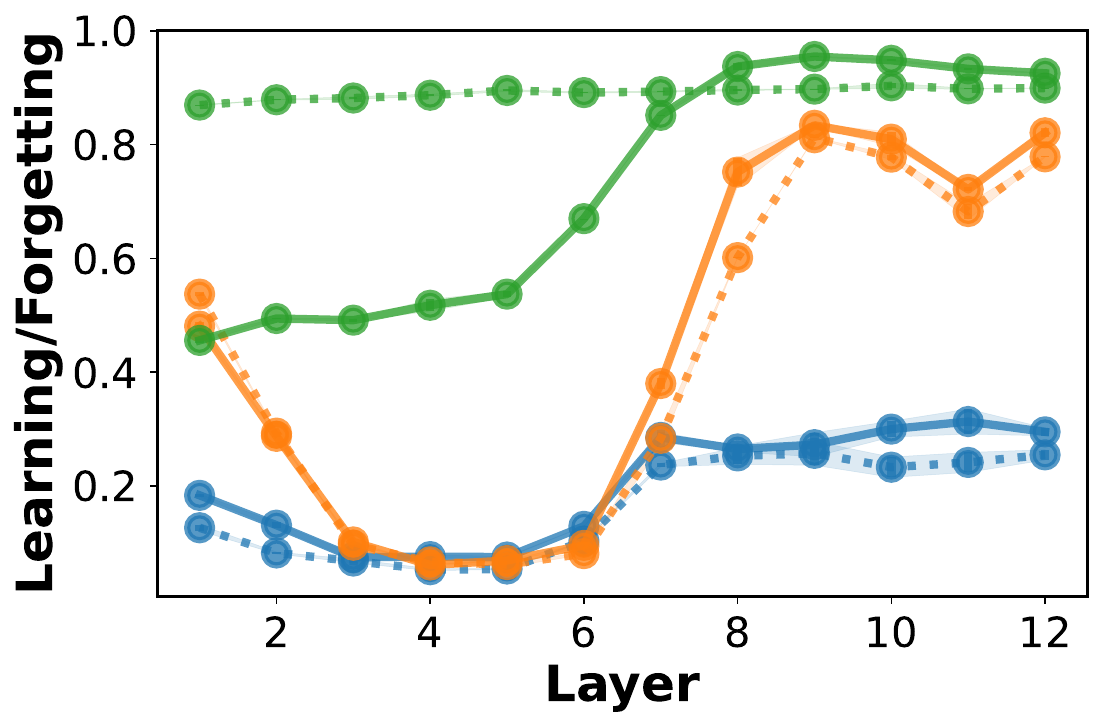}
        \caption{CIFAR10}
        \label{fig:cifar10_diff_pretrainings}
    \end{subfigure}
    \begin{subfigure}[b]{0.35\textwidth}
        
        \includegraphics[width=1\textwidth]{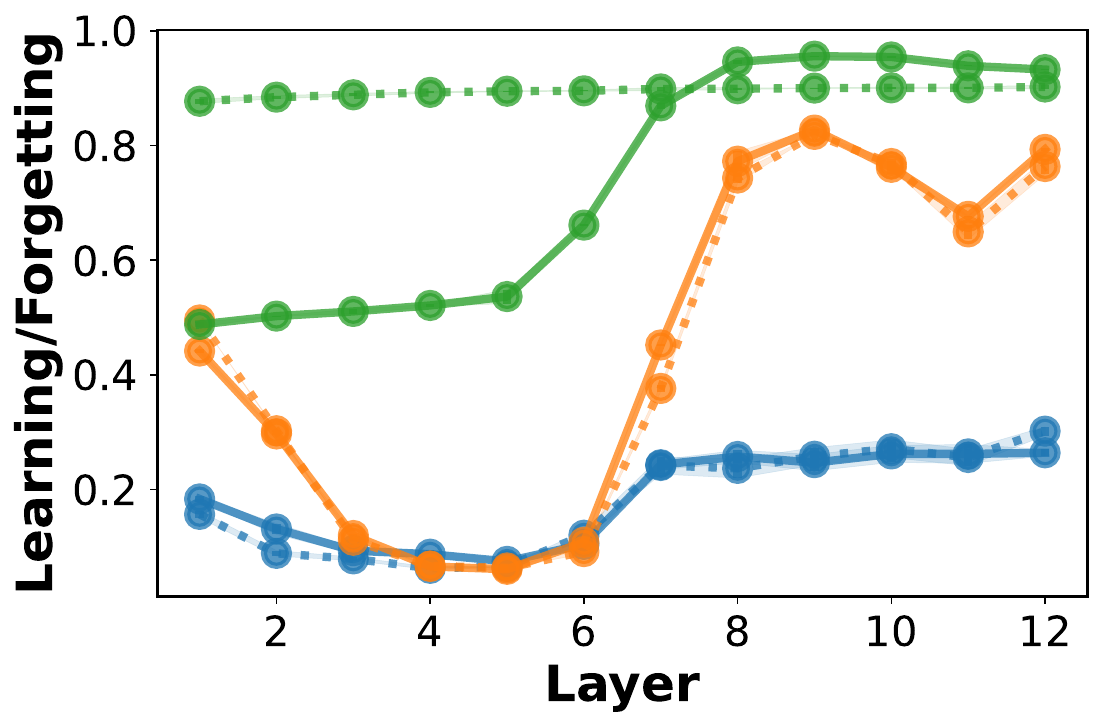}
        \caption{CIFAR100}
        \label{fig:cifar100_diff_pretrainings}
    \end{subfigure}
    \raisebox{0.99\height}{
    \begin{subfigure}[b]{0.25\textwidth}
     
        \includegraphics[width=1.1\textwidth]{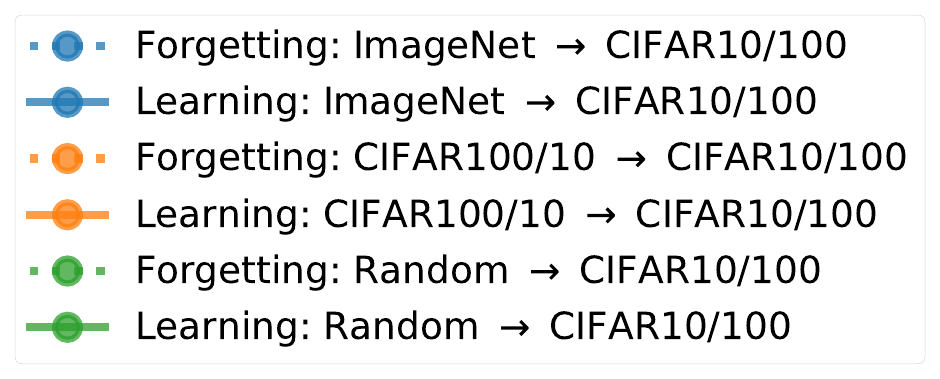}

    \end{subfigure}
    }
    \caption{\textbf{[Pretraining dataset affects \texttt{learning} and \texttt{forgetting}]} \texttt{learning} and \texttt{forgetting} across layers for a model finetuned for CIFAR10 (left) and CIFAR100 (right) classification, starting from different pretrained weights. The models are pretrained on CIFAR100/10 (orange), ImageNet (blue), or trained from scratch (random initialization, shown in green). For CIFAR10 (left), the orange line shows results for pretraining on CIFAR100, while for CIFAR100 (right), it shows results for pretraining on CIFAR10.  Finetuning a model pretrained on \textit{some} data leads to a lower \texttt{learning} and \texttt{forgetting} of invariances in early layers, whereas training from scratch leads to much higher learning (and forgetting), even in early layers. We also see that both pretraining datasets instill reusable invariances in early-to-mid layers. Further, ImageNet pretraining leads to useful invariances even in later layers, thus indicating why such pretraining is widely used as a recipe for a variety of computer vision tasks.}
  \label{fig:pre_learning_forgetting}
\end{figure}

\begin{figure}[t]
  \centering

    \begin{subfigure}[b]{0.35\textwidth}
    
        \includegraphics[width=1\textwidth]{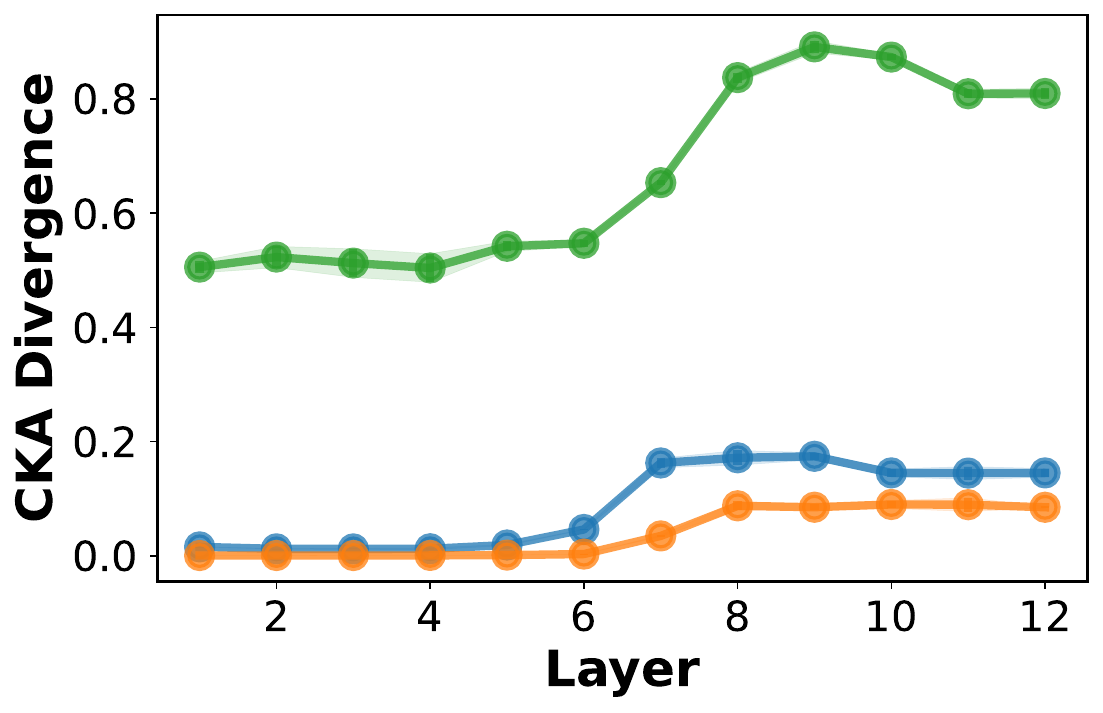}
        \caption{CIFAR10}
        \label{fig:cifar10_diff_pretrainings_cka}
    \end{subfigure}
    \begin{subfigure}[b]{0.35\textwidth}
        
        \includegraphics[width=1\textwidth]{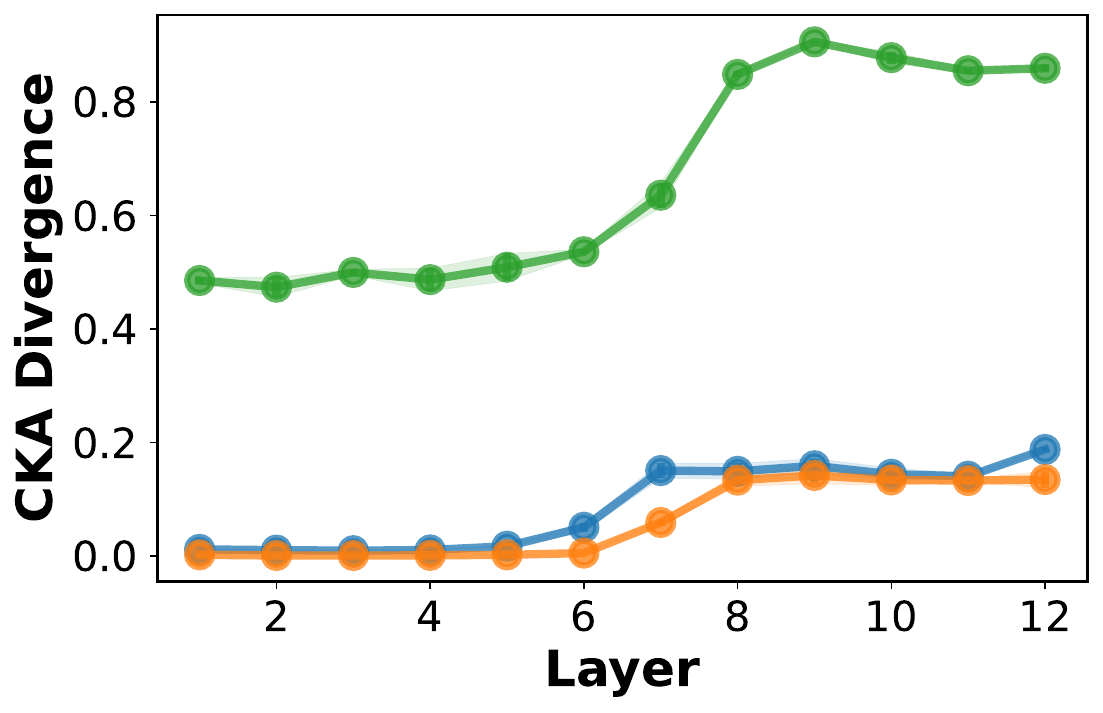}
        \caption{CIFAR100}
        \label{fig:cifar100_diff_pretrainings_cka}
    \end{subfigure}
    \raisebox{1.4\height}{\begin{subfigure}[b]{0.25\textwidth}
        \includegraphics[width=1\textwidth]{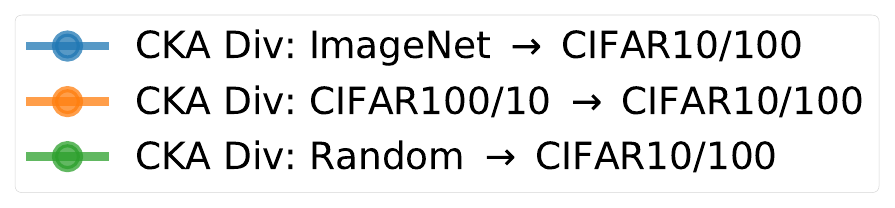}
        \label{fig:cifar100_diff_pretrainings_cka_legend}
    \end{subfigure}}
    \caption{\textbf{[\texttt{cka\:divergence} does not capture changing nature of invariances]} \texttt{cka\:divergence} across layers for a model finetuned for CIFAR10 (left) and CIFAR100 (right) classification, starting from different pretrained weights. 
    Finetuning a model pretrained on \textit{some} data leads to almost no \texttt{cka\:divergence} in early layers, and only small values in later layers. However, training from scratch leads to a much higher \texttt{cka\:divergence}, even in early layers. Interestingly, contrary to the trend shown for \texttt{learning} and \texttt{forgetting} in Fig~\ref{fig:pre_learning_forgetting}, we see that ImageNet pretraining leads to higher divergence than pertaining on the respective CIFAR dataset, as shown by the blue line being higher than orange in both plots.}
  \label{fig:pre_cka_div}
\end{figure}

To analyze the effect of a pretrained dataset on the forgotten and learned invariances during finetuning, we analyze models that are initialized using different weights, and are then all finetuned for the same task (classification) and dataset (CIFAR10 in Fig~\ref{fig:cifar10_diff_pretrainings} and CIFAR100 in Fig~\ref{fig:cifar100_diff_pretrainings}). For each finetuning dataset, we consider 2 models with initial weights that are obtained by (pre) training on ImageNet, or on either CIFAR100 or CIFAR10 depending on the finetuning dataset. As a baseline, we additionally consider a third model that's trained from scratch for each dataset (\ie\ starting from a random weight initialization). We present the values of the \texttt{forgetting} and \texttt{learning} metrics for all models in Fig~\ref{fig:pre_learning_forgetting} and \texttt{cka\:divergence} in Fig~\ref{fig:pre_cka_div}.

\paragraph{Pretraining instills reusable invariances in early layers.}
We observe that the finetuned models that start from pretrained models exhibit a substantially lower \texttt{learning} and \texttt{forgetting} in the early layers (1-6), than the model trained from scratch (green lines in Fig~\ref{fig:pre_learning_forgetting}).
This suggests that pretraining instills certain invariances during pretraining that can be reused in the finetuning task. This finding is consistent with other work showing that earlier layers of the pretrained model preserve more general knowledge that is still useful during finetuning~\citep{ramasesh2020anatomy,zeiler2014visualizing}.

\paragraph{ImageNet pretraining leads to more useful invariances even in later layers.} In Fig~\ref{fig:pre_learning_forgetting} we observe that even for later layers (\ie~7-12), the learning and forgetting values for the model pretrained on ImageNet are significantly lower than those for CIFAR10/100 pretraining or training from scratch. Our observation aligns with decades of empirical results that have found ImageNet pretraining to be an effective strategy for a variety of computer vision tasks~\cite{girshick2014rich,kornblith2019better,long2015fully}.

\paragraph{Training from scratch requires higher learning and forgetting across all layers} For both finetuning on CIFAR10 and CIFAR100, we see that when learning from scratch (\ie\~random weight initialization; shown in green in Fig~\ref{fig:pre_learning_forgetting}) \texttt{learning} and \texttt{forgetting} both have higher values than pretrained models, across all layers. This aligns well with well-studied gains of pretraining in prior works~\cite{donahue2014decaf,erhan2010does}.

\paragraph{Representation similarity does not faithfully indicate the effects of pretraining.} In Fig~\ref{fig:pre_cka_div} we show \texttt{cka\:divergence} for the same setting evaluated in Fig~\ref{fig:pre_learning_forgetting}. We observe that contrary to our finding about \texttt{learning} and \texttt{forgetting}, \texttt{cka\:divergence} between pretrained and finetuned model shows higher values when using ImageNet pretrained weights than the corresponding CIFAR pretrained weights. Similar to observations of~\cite{nanda2022measuring}, CKA does not capture the nature of changing invariances and thus can give incomplete information about the effects of pretraining.

\subsection{Compression and Expansion Analysis}
\label{subsec:copression_and_expansion_analysis} 

Thus far we have focused on examining how the invariances learned by a specific layer in a pretrained model relate to the corresponding layer in the finetuned model. However, this may not capture all possible effects of finetuning as invariances in some pretrained layers may have migrated to other layers in a finetuned model due to task-dependent requirements. Here, we use the \texttt{Invariance\:Flow\:Matrix} to examine whether layer-wise invariances in the pretrained model closely correspond to the respective layer in the finetuned model, or whether there are better-matching layers.

\begin{figure}[t]
\centering
    \begin{subfigure}{0.35\textwidth}
    
        \includegraphics[width=\textwidth]{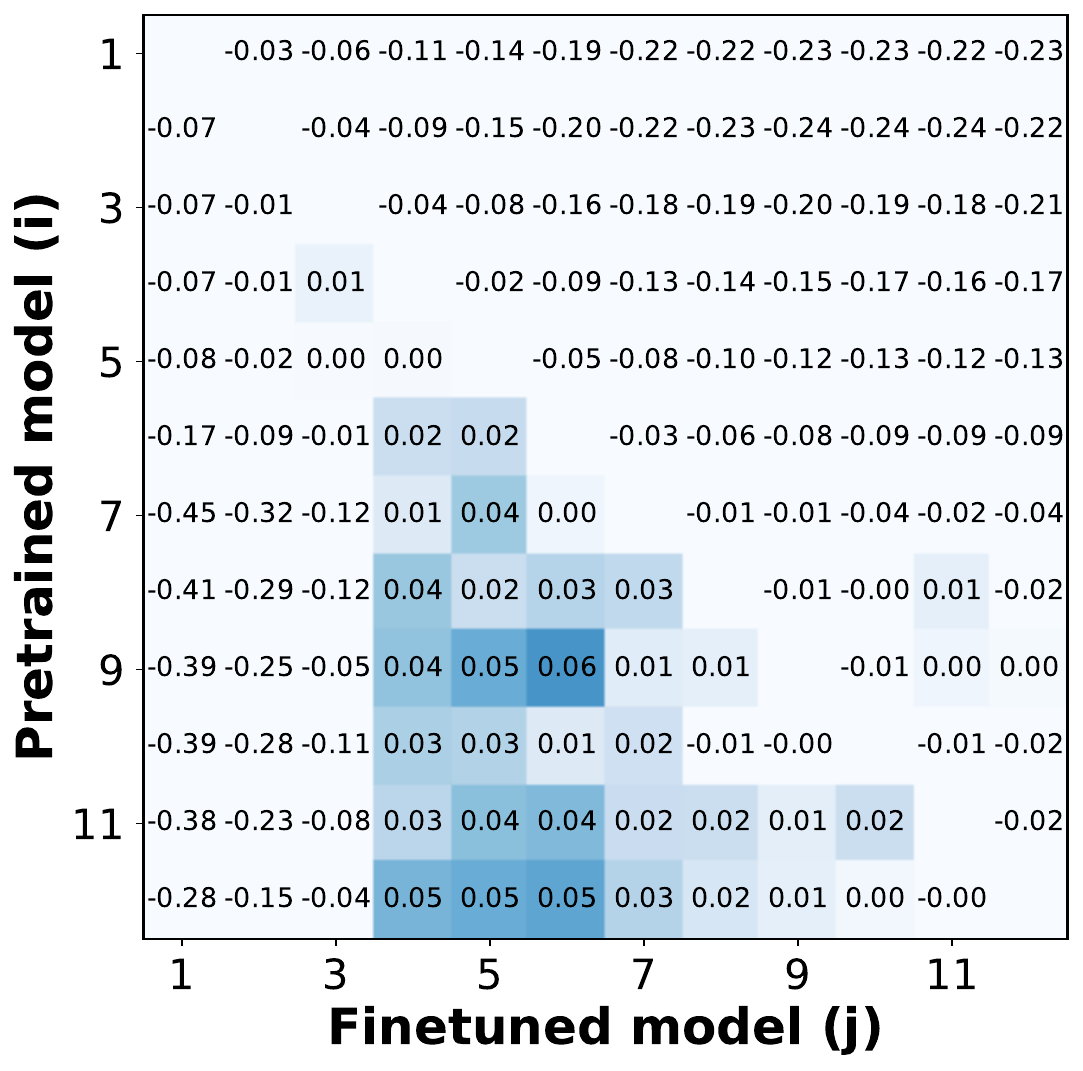}
        \caption{Classification Task}
        \label{fig:comp_class_cifar10}
    \end{subfigure}
    \hspace{0.15\textwidth}
   \begin{subfigure}{0.35\textwidth}
   
        \includegraphics[width=\textwidth]{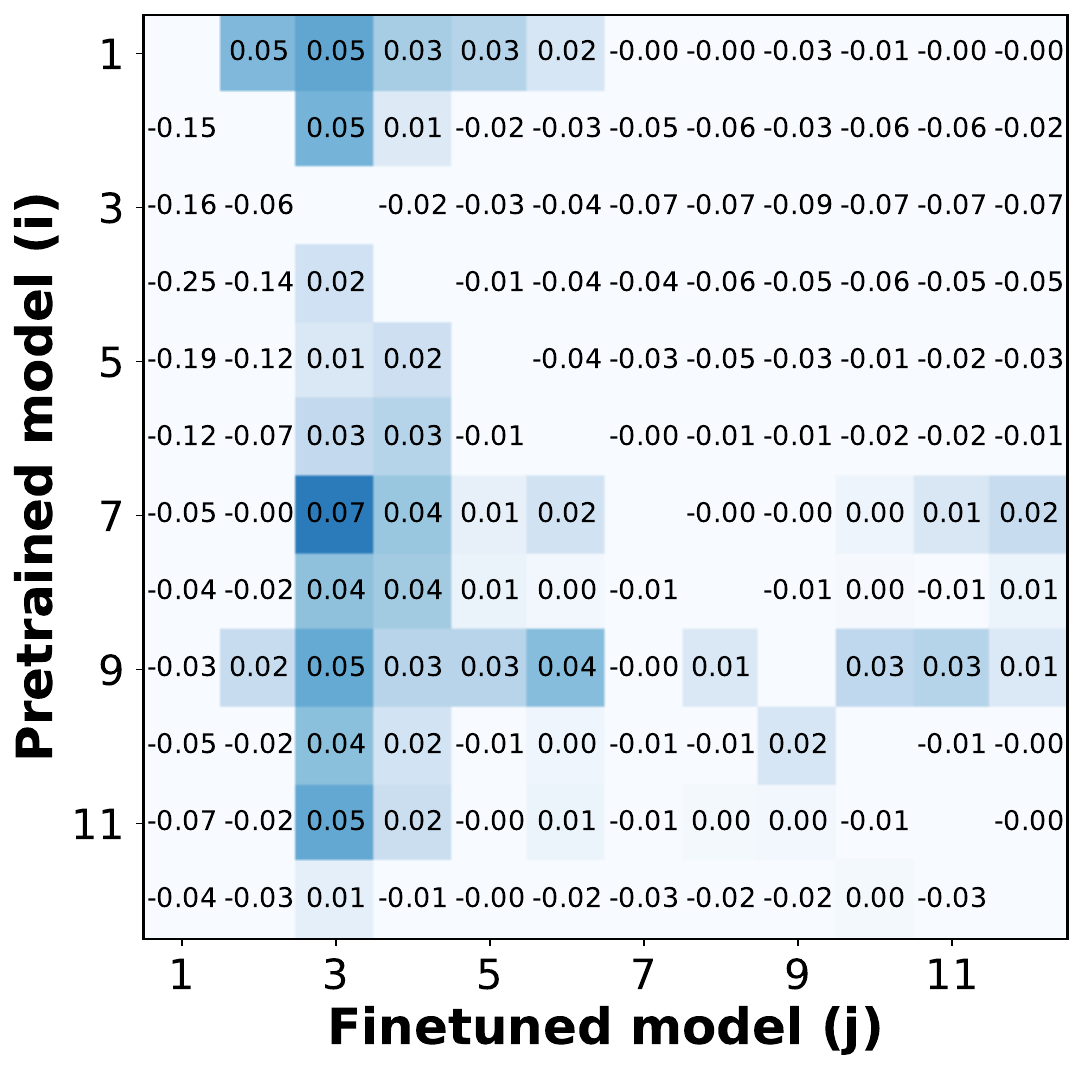}
        \caption{Reconstruction Task}
        \label{fig:comp_recon_cifar10}
    \end{subfigure}
   \caption{Invariance Flow Matrix for the finetuned model on classification and reconstruction task of CIFAR10, pretrained on ImageNet. For the classification task we observe a compression of the pretrained invariances. In particular, layers $6-12$ in the pretrained model correspond more closely to layers $4-8$ in the finetuned model. For the reconstruction task we observe both a compression and expansion of the pretrained invariances. In particular, we observe that layers $6-12$ in the pretrained model correspond more closely to layers $3-6$ in the finetuned model, which indicates compression. In contrast, layers $1-2$ in the pretrained model correspond more closely to layers $2-6$ in the finetuned model, which indicates expansion.}
\end{figure}

\paragraph{Finetuning compresses invariances.} In Figures \ref{fig:comp_class_cifar10} and \ref{fig:comp_recon_cifar10}, we present the \texttt{Invariance\:Flow\:Matrix} for the model pretrained on ImageNet and finetuned for classification and reconstruction of CIFAR10 respectively. For both tasks, we observe a compression of the pretrained invariances, visible in the lower left part of the matrix. In particular, layers $6-12$ in the pretrained model correspond more closely to layers $4-8$ in the model finetuned for classification and layers $4-11$ correspond more closely to layers $3-6$ in the model finetuned for reconstruction. 
 This suggests that finetuning  compresses some invariances from the pretrained model to earlier layers of the finetuned model. This is possibly a mechanism that allows for more capacity in later layers to support learning new invariances that are needed for the finetuning task and dataset. We observe similar results for the classification and reconstruction of the CIFAR100 dataset (see Appendix Figures \ref{fig:comp_class_cifar100} and \ref{fig:comp_recon_cifar100}).

\paragraph{Transfering to lower-level tasks expands early-layer invariances.}
In the upper right triangles of Figures \ref{fig:comp_class_cifar10} and \ref{fig:comp_recon_cifar10}, we can observe any possible expansion effects of finetuning as discussed in Section \ref{sec:methods}. Interestingly this effect is only observable when the model is finetuned on the reconstruction CIFAR10/CIFAR100 (see Appendix Figures \ref{fig:comp_class_cifar100} and \ref{fig:comp_recon_cifar100} for similar CIFAR100 results). This observation is consistent with our expectations. Reconstruction is a lower-level task than classification: to reconstruct an image, more local information may be useful, and usually this kind of information is stored in earlier layers. The \texttt{Invariance\:Flow\:Matrix} shows that low-level invariances present in earlier layers of the pretrained model are now useful in deeper layers of finetuned model to perform reconstruction.

\subsection{Forgetting and Learning Dynamics}
\label{subsec:training_dynamics} 

So far we analyzed the \texttt{learning} and \texttt{forgetting} metrics across different layers to quantify the difference between the pretrained and the finetuned model. However, a neural network changes gradually during training, increasing its accuracy on the finetuning task. In this section, we examine the evolution of the proposed metrics during finetuning.

\begin{figure}[t]

\raggedright
    \hspace{0.13\textwidth}
    \begin{subfigure}[t]{0.35\textwidth}
        
        \vskip 0pt
        \includegraphics[width=\textwidth,]{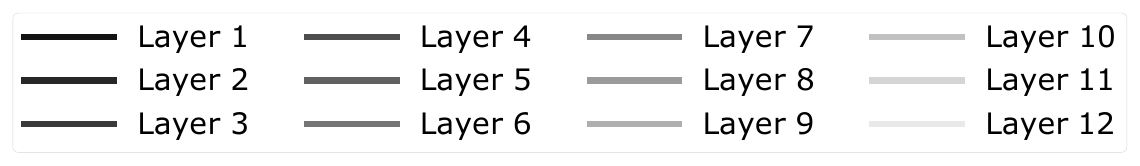}
        
    \end{subfigure}
    
\centering
    \begin{subfigure}[t]{0.28\textwidth}
        \vskip 0pt
        
        \includegraphics[width=\textwidth]{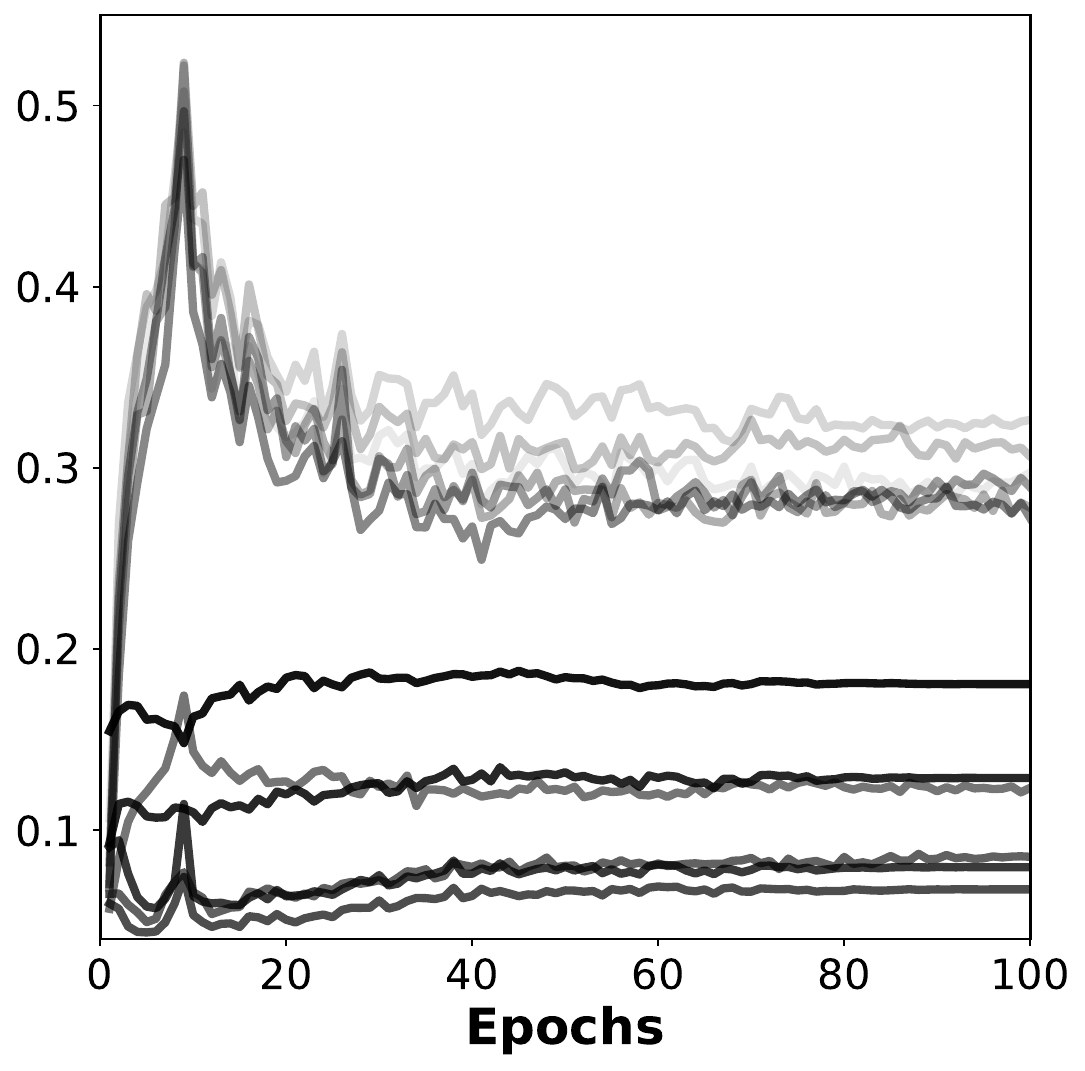}
        \caption{\texttt{learning} during finetuning}
        \label{fig:during_learn}
    \end{subfigure}
   \begin{subfigure}[t]{0.28\textwidth}
        \vskip 0pt
        
        \includegraphics[width=\textwidth]{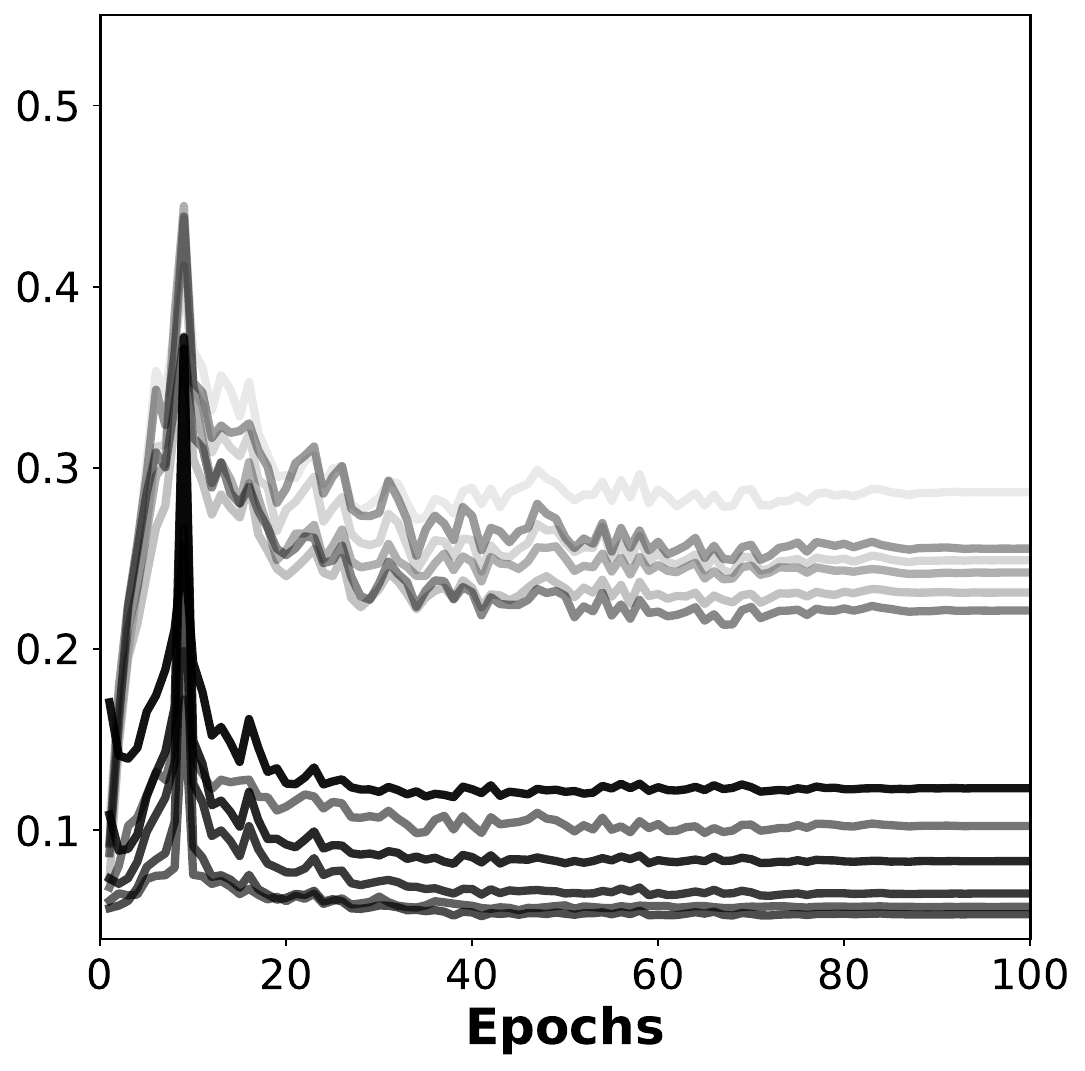}
        \caption{\texttt{forgetting} during finetuning}
        \label{fig:during_forget}
    \end{subfigure}
    \begin{subfigure}[t]{0.42\textwidth}
        \vskip 0pt
        
        \includegraphics[width=\textwidth]{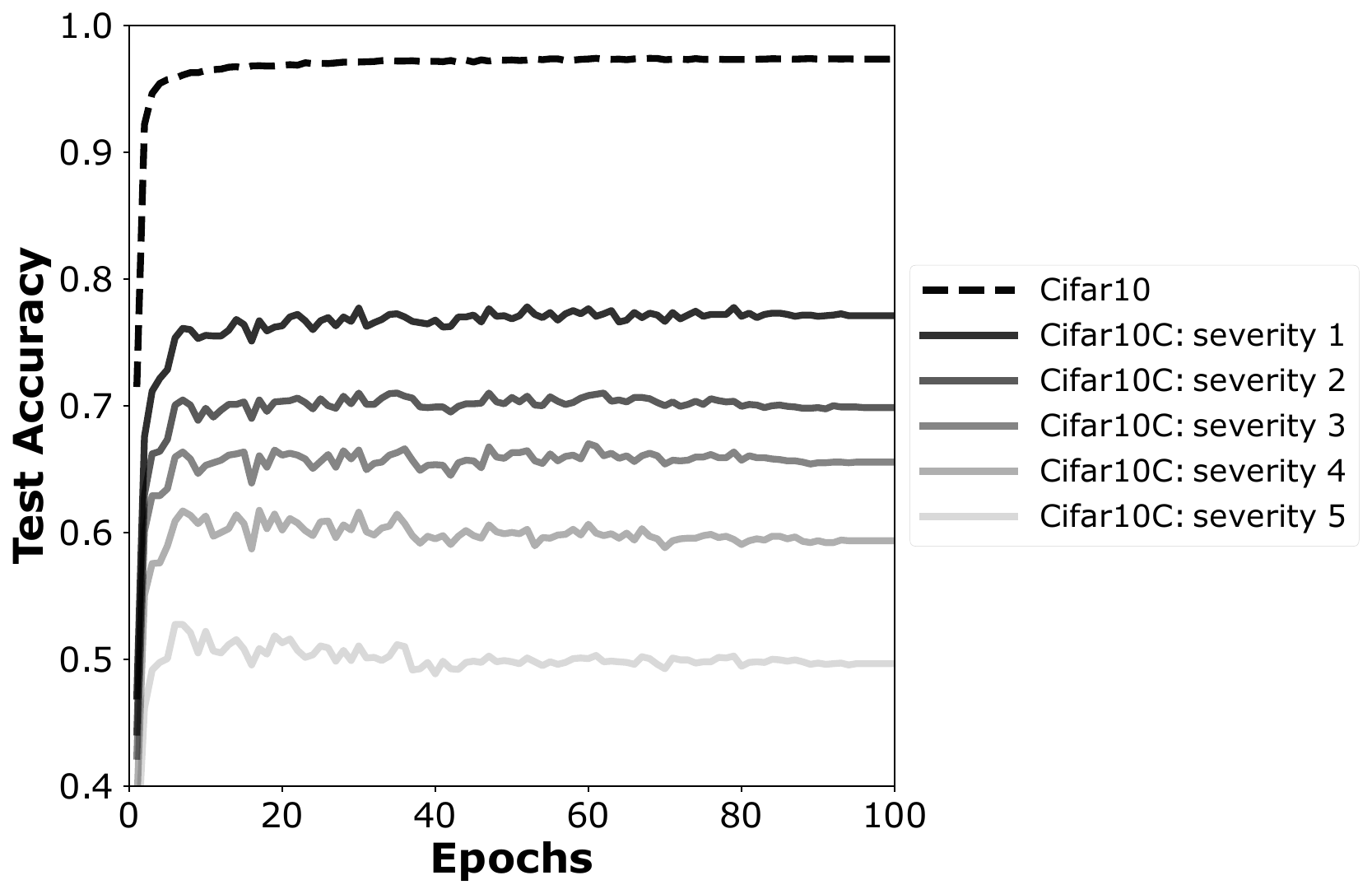}
        \caption{Accuracies during finetuning}
        \label{fig:during_acc}
    \end{subfigure}
    \captionsetup[subfigure]{labelformat=empty}

   \caption{\texttt{(a-b)} \texttt{learning} and \texttt{forgetting} during finetuning on classification task of CIFAR10 of a model pretrained on ImageNet. We observe a peak of the two metrics in earlier epochs. Only the \texttt{learning} of earlier layer does not exhibit a peak.\texttt{(c)} Test accuracy during finetuning on classification task of CIFAR10 of a model pretrained on ImageNet. Different lines correspond to different corruption level of the original CIFAR10 test set. While the accuracy on the standard test set increase, the accuracy on corrupted dataset does not always increase, in particular for dataset with high level of corruption(\ie\ severity 3, 4, 5). In these cases the accuracy has a peak on earlier epochs.}
\end{figure}

\paragraph{Learning and Forgetting do not increase monotonically.}
Usually during training the accuracy on the test set increases gradually and the model becomes increasingly capable of performing the task. In Figure \ref{fig:during_acc} we can clearly observe this trend (\ie\ dotted line). However, analyzing the \texttt{learning} and \texttt{forgetting} during training and across different layers in Figure \ref{fig:during_learn} and \ref{fig:during_forget} we can observe that the two metrics do not increase monotonically. This is counter-intuitive from a behavioural perspective since one could expect an increasing \texttt{learning} and \texttt{forgetting} as the model continues to perform better during training. In particular later layers exhibit a peak in earlier epochs. This means that in earlier epochs the model has learned and forgotten invariances, as a result the model diverges more with respect to the pretrained model. This observation is confirmed by the \texttt{cka\:divergence} in Appendix Figures \ref{fig:during_cka}, \ref{fig:during_cka_cifar100} and it is true also for the CIFAR100 dataset (Appendix Figures \ref{fig:during_learn_cifar100}, \ref{fig:during_forget_cifar100}, \ref{fig:during_acc_cifar100},).

\paragraph{Variability in forgetting across layers could reveal more robust models.}
A question that naturally arises upon observing the different dynamics of \texttt{learning} and \texttt{forgetting} with respect to the test accuracy is whether these metrics could reveal additional properties of the model, such as robustness of the model to data corruption.
In Figure \ref{fig:during_acc} we report the accuracy of the model on corrupted CIFAR10 test sets with different levels of corruption. For this purpose, we use the benchmark proposed by \cite{croce2020robustbench}. In correspondence with the peak of \texttt{learning} and \texttt{forgetting}, showed in Figures \ref{fig:during_learn}, \ref{fig:during_forget}, we observe higher robustness of the model in particular on the dataset with higher levels of corruption (\ie\ severity 3, 4, 5). We hypothesize that the interaction between learning and forgetting of different layers could be correlated with robustness accuracy. To test whether there is a notable relationship between $\texttt{learning}/$\texttt{forgetting} and the robustness accuracy, we compute the Pearson correlation between these metrics across training epochs (details in Appendix \ref{app:corr_robust}). We observe a strong correlation (0.78 as correlation value, p-value 3.52e-18 that passes the Bonferroni correction used to take into account the multiple possible hypotheses) on average across the robustness accuracy on corrupted CIFAR10 test sets and the standard deviation of \texttt{forgetting} across layers $2-12$. The correlation is high even for the corrupted CIFAR100 test sets (0.877 as correlation value, p-value 2.51e-16 that passes the Bonferroni correction). This suggests that when \texttt{forgetting} varies a lot across layers, there is also more robustness. 
We repeated this test choosing the best aggregate metrics for both \texttt{cka\:divergence} and  $subspace\:similarity$ \citep{ramasesh2022effect}, and we have found a weaker correlation between \texttt{cka\:divergence} and robustness accuracy across corrupted CIFAR10 test sets (std.\ dev.\ of cka divergence values, layers 1-8, 0.69) and CIFAR100 test sets (0.63) and between $subspace\:similarity$ and robustness accuracy (0.44 CIFAR10c, 0.69 CIFAR100c).

We also note that both the standard deviation of \texttt{forgetting} across layers (Figure~\ref{fig:during_forget}) and average accuracies on corrupted datasets (Figure~\ref{fig:during_acc}) stabilize to constant values towards the end of training. 
Hence, a valid concern regarding the correlation analysis mentioned earlier is whether the high correlation value stems from the trend of stabilization observed towards the end of the training, rather than the trend of coinciding peaks in both the standard deviation of forgetting and average accuracy values during the initial phase of training.
The latter is especially interesting for practitioners seeking to utilize higher values of the standard deviation of \texttt{forgetting} as an early stopping indicator for robustness.
Therefore, we conduct further experiments to disentangle the contribution of the early epochs by computing the correlation between average accuracies up to an epoch $n$ with the standard deviation of forgetting up to epoch $n$. 
Thus, if the correlation is still strong considering only the initial epochs, then this suggests that the higher overall correlation value is not solely attributable to the later epochs.
Interestingly, in Figure~\ref{fig:corr_analysis} (Appendix), we observe precisely that, as the correlation up to early epochs is not only comparable but even higher than up to the epochs towards the end of training.  

\begin{figure}[t]

  \centering
  \includegraphics[width=0.5\columnwidth]{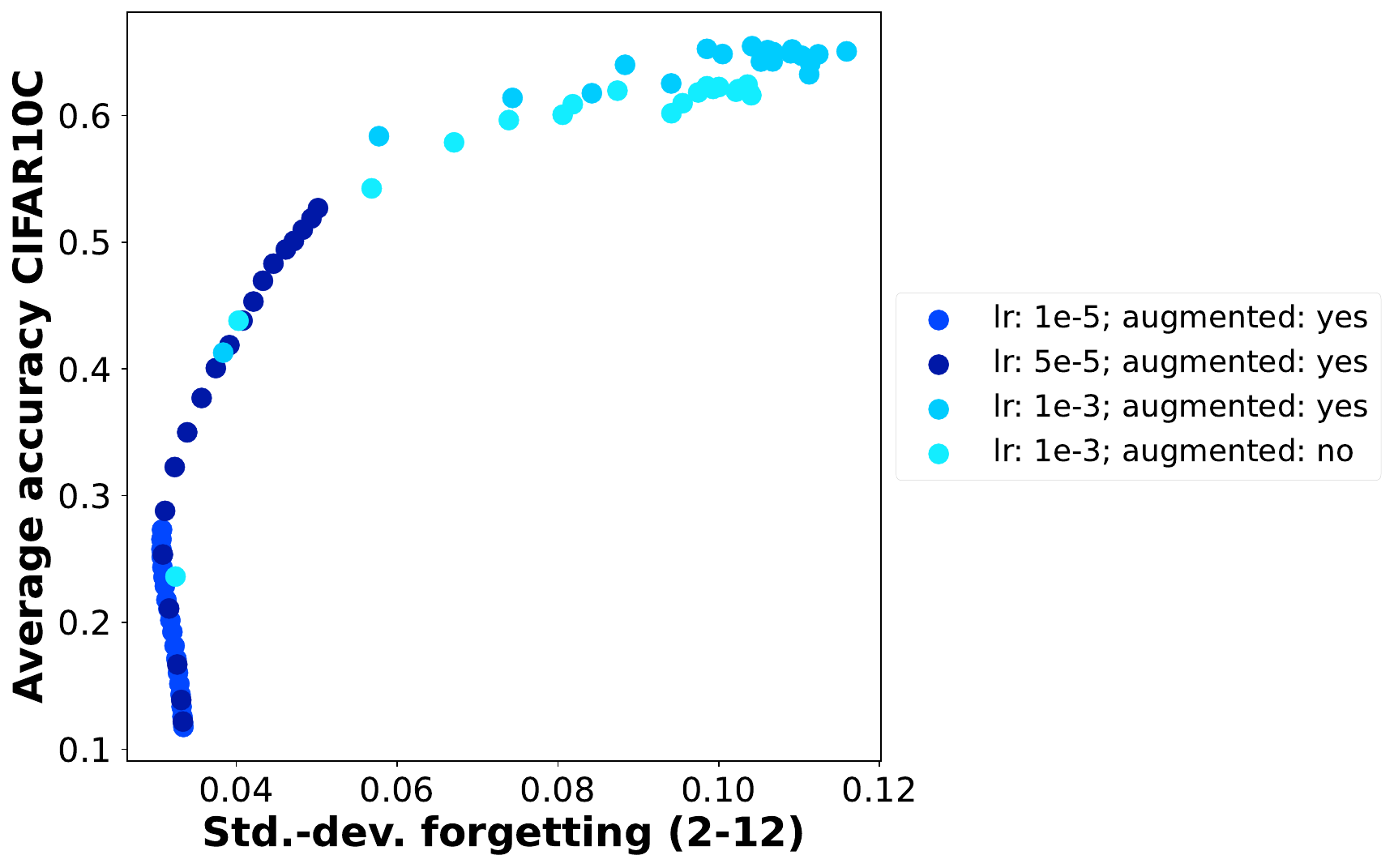}
  \caption{Average accuracy on CIFAR10C datasets and standard deviation of \texttt{forgetting} across layers 2-12 of 20 epochs of different models while training on classification of CIFAR10 with $sgd$ optimizer. Higher forgetting variability across layers corresponds to higher accuracy on corrupted datasets.}
  \label{fig:acc_forg_sgd}

\end{figure}

The analysis of the correlation reported in the previous paragraph was conducted for a single model. To validate our findings beyond this model setting, we observed the corresponding relationship when varying several hyperparameters. We report in Figure \ref{fig:acc_forg_sgd} the average accuracy on the corrupted CIFAR10 datasets and the standard deviation of \texttt{forgetting} across layers 2-12 during the first 20 epochs of models trained on CIFAR10 classification with $sgd$ as optimizer with different hyperparameter settings. The different models have similar trends, even with different learning rates and the usage of augmentations. Interestingly, when using $adam$ as optimizer we did not find a clear trend as with $sgd$ and higher standard deviation of \texttt{forgetting} does not coincide with higher accuracy on corrupted datasets (Figure \ref{fig:acc_forg}). We leave a deeper exploration into the effect of different optimizers on the relationship between model robustness and standard deviation of forgetting across layers for future work.

\section{Conclusion}
This work examined the relationship between pretrained vision transformers(ViT) and the corresponding finetuned versions on several benchmark datasets and tasks. We presented new metrics that specifically investigate the degree to which invariances learned by a pretrained model are learned or forgotten during finetuning (Section \ref{sec:methods}). Using these metrics, we presented empirical results on the effect of the finetuning task and the pretraining dataset on the invariances (Section \ref{subsec:learning_and_forgetting_invariances}). We further showed that invariances from deeper pretrained layers are compressed towards shallower layers during finetuning, which may be a mechanism that allows for more capacity in later layers to support learning new invariances that are needed for the finetuning task and dataset (Section \ref{subsec:copression_and_expansion_analysis}). Analyzing the \texttt{learning} and \texttt{forgetting} dynamics during finetuning (Section \ref{subsec:training_dynamics}), we show that they do not increase monotonically as was expected and we revealed a strong correlation between these metrics, and in particular the standard deviation of \texttt{forgetting} across layers, and the robustness of the model. This correlation becomes even stronger when fewer epochs are considered, making this measures particularly useful to analyze the robustness of the model during training. Together, these findings contribute to understanding some of the reasons for the successes of pretrained models and the changes that a pretrained model undergoes when finetuned on a downstream task. 

The aim of our work was to provide a deeper understanding of changes in different layers during finetuning of ViTs. We offered a novel perspective on finetuning by analyzing model changes through the lens of shared invariances. There is already a rich body of ongoing studies that introduce better strategies for finetuning \citep{kumar2022fine,lee2022surgical,
  evci2022head2toe}. These studies show that early layers features can be leveraged for better transfer performance. Our work instead aims to shed light on \textit{why} certain approaches work, by showing that early layers tend to learn transferable invariances. Our analysis can inspire future work to design even more effective architectures and finetuning strategies.

\subsubsection*{Acknowledgments}
The authors would like to thank Camila Kolling and Till Speicher for helpful feedback on an earlier version of this manuscript. GM was supported by the CS@max planck graduate center. VN was supported in part by an ERC Advanced Grant “Foundations for Fair Social Computing” (no. 789373), NSF CAREER Award IIS-1846237, NSF D-ISN Award \#2039862, NSF Award CCF-1852352, NIH R01 Award NLM013039-01, NIST MSE Award \#20126334, DARPA GARD \#HR00112020007, DoD WHS Award \#HQ003420F0035, ARPA-E Award \#4334192. MT was supported in part by the German Research Foundation (DFG) - DFG Research Unit FOR 5368.

\bibliography{collas2023_conference}
\bibliographystyle{collas2023_conference}

\newpage
\appendix
\section{Appendix}
\section{Training details}
\label{app:training_details}
For our experiment we use a ViT model with patch size of 32x32 and image
resolution at 224x224. The pretrained model on ImageNet is provided by \citet{rw2019timm}. We train the other models for 100 epochs with $learning\:rate=0.001$, $momentum=0.9$ and $weight\:decay=0.0001$. We use a cosine scheduler for the learning rate and the Stochastic Gradient Descent as optimizer. We use the \texttt{transformers} library from Hugging Face\cite{wolf2020transformers} to train the model and log training results.
In table \ref{tab:accuracy}
we report the accuracy values on the test set for the model we use.
\begin{table}[h]
\caption{Accuracy values ViT models}
\label{tab:accuracy}
\vskip 0.15in
\begin{center}
\begin{small}
\begin{sc}

\begin{tabular}{|l|l|l|}
\hline
\textbf{Model}                      & \textbf{Accuracy} & \textbf{Pretraining Accuracy} \\ \hline
Pretrain ImageNet; finetune CIFAR10 & 0.97        &  0.51                        \\ \hline
Pretrain ImageNet; finetune CIFAR100 & 0.86        & 0.51                            \\ \hline
Pretrain CIFAR100; finetune CIFAR10 & 0.99              & 0.92                       \\ \hline
CIFAR10 from scratch                    & 0.98              & --                             \\ \hline
Pretrain CIFAR10; finetune CIFAR100                    & 0.91              & 0.98                           \\ \hline
CIFAR100 from scratch                    & 0.92              & --                             \\ \hline
Pretrain ImageNet; finetune Eurosat & 0.97        & 0.51                            \\ \hline
Pretrain ImageNet; finetune OxfordIIIT Pet & 0.80        & 0.51                            \\ \hline
\end{tabular}
\end{sc}
\end{small}
\end{center}
\vskip -0.1in
\end{table}

\newpage
\section{Effect of finetuning/pretraining tasks: additional results}
\label{app:task_cifar100}

\begin{figure}[ht]
\centering
    \begin{subfigure}[b]{0.30\textwidth}
        
        \includegraphics[width=1\textwidth]{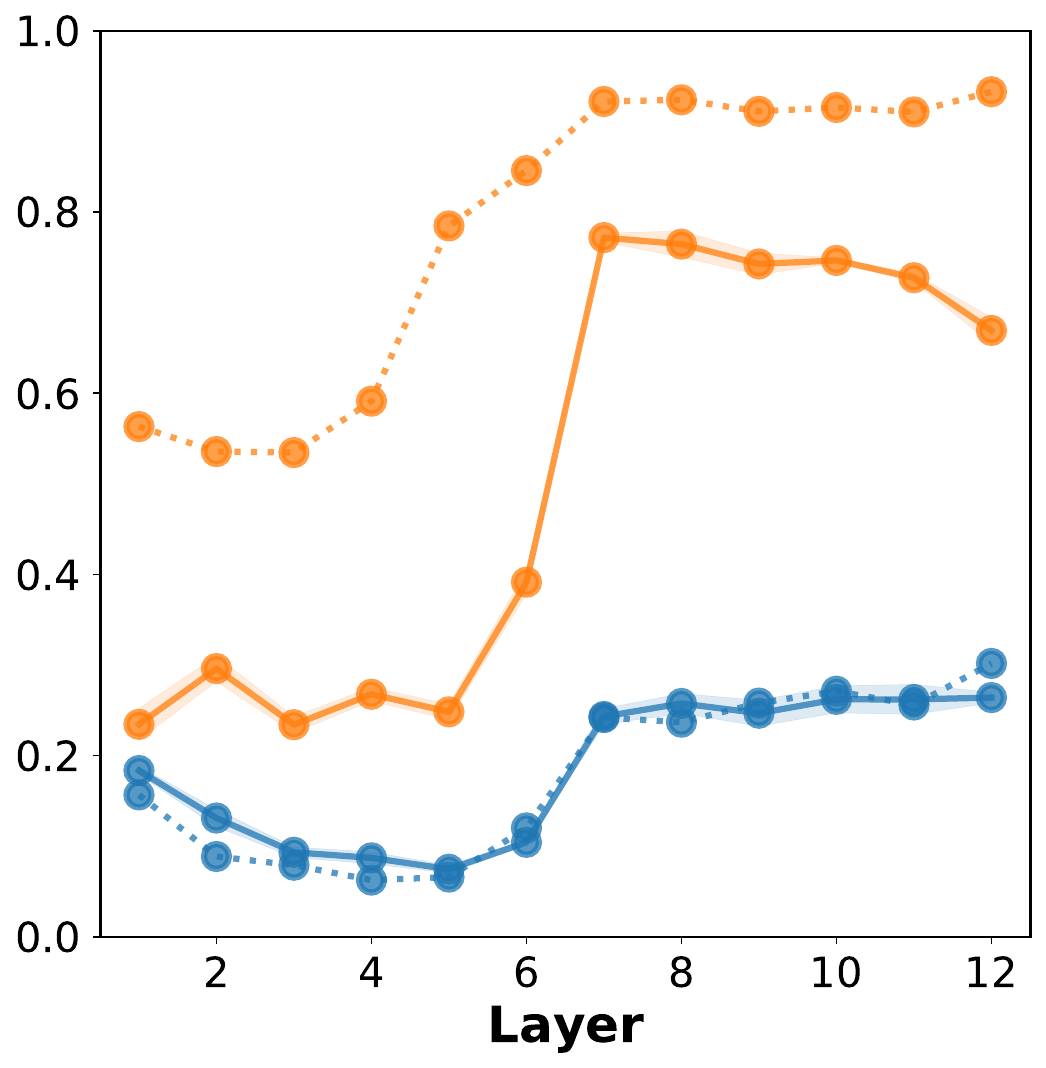}
        \caption{Learning and forgetting CIFAR100}
        \label{fig:learning_forgetting_cifar100}
    \end{subfigure}
    \captionsetup[subfigure]{labelformat=empty}
   \raisebox{3.5\height}{\begin{subfigure}[b]{0.18\textwidth}
        \includegraphics[width=1\textwidth]{images/learn_forget_legend.pdf}
    \end{subfigure}
    }
    \captionsetup[subfigure]{labelformat=parens}
    \begin{subfigure}[b]{0.30\textwidth}
        \includegraphics[width=1\textwidth]
        {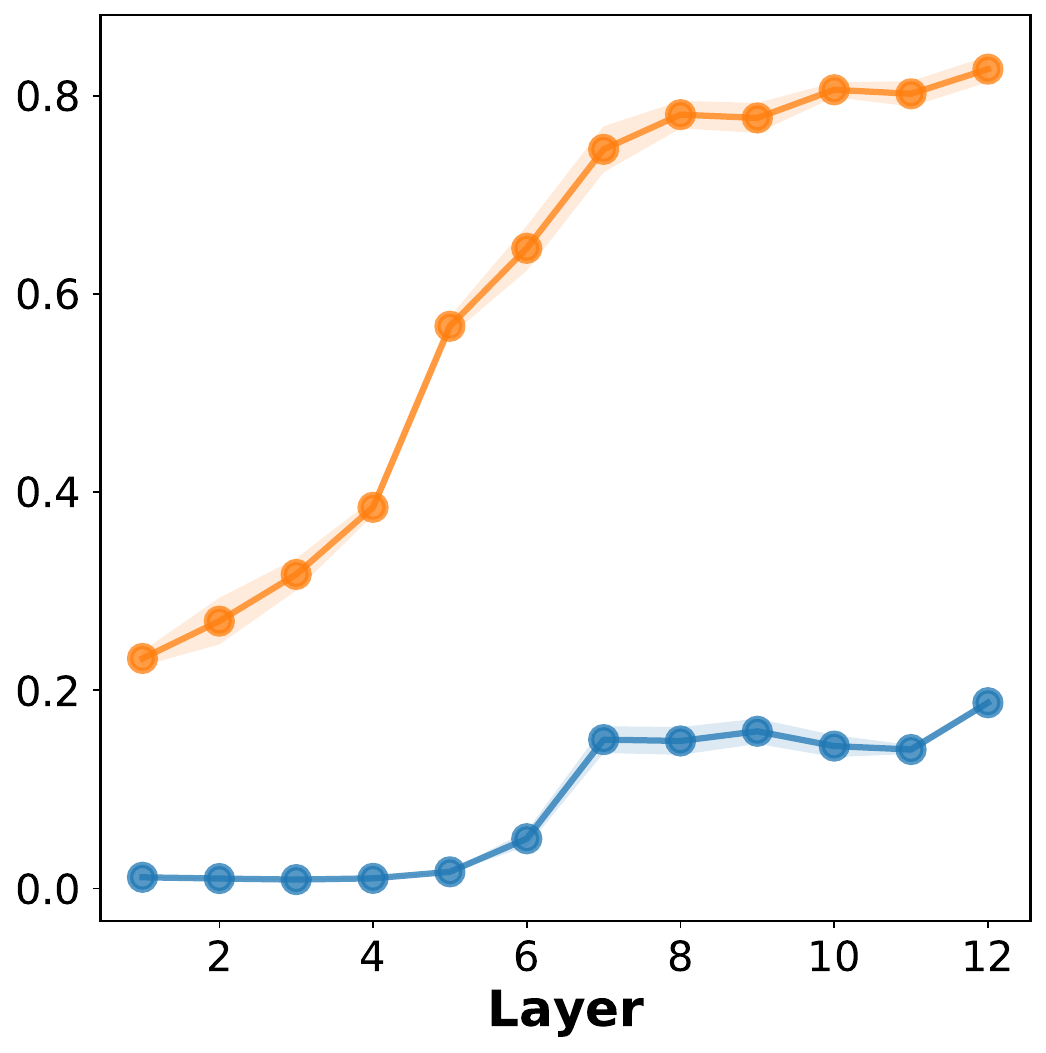}
        \caption{CKA divergence CIFAR100}
        \label{fig:cka_divergence_cifar100}
    \end{subfigure}
    \captionsetup[subfigure]{labelformat=empty}
   \raisebox{7\height}{\begin{subfigure}[b]{0.18\textwidth}
        \includegraphics[height=0.17\textwidth]{images/cka_legend.pdf}
    \end{subfigure}
    }
   \caption{\texttt{learning}, \texttt{forgetting} and \texttt{cka\:divergence} values for ViT model pretrained on ImageNet and finetuned on CIFAR100 on reconstruction and classification task.}
\end{figure}

\begin{figure}[ht]
\centering
    \begin{subfigure}[b]{0.30\textwidth}
        
        \includegraphics[width=1\textwidth]{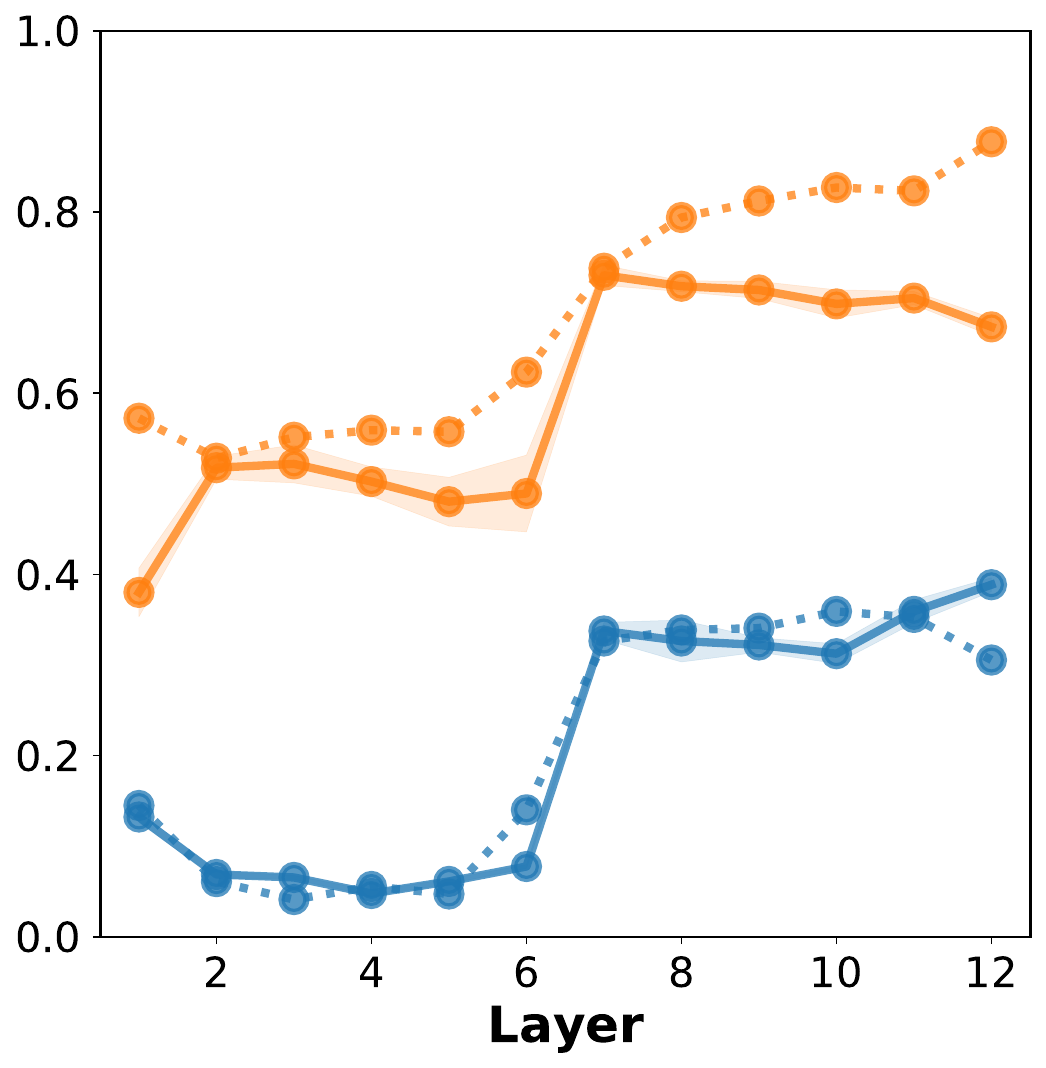}
        \caption{Learning and forgetting EuroSAT}
        \label{fig:learning_forgetting_eurosat}
    \end{subfigure}
    \captionsetup[subfigure]{labelformat=empty}
   \raisebox{3.5\height}{\begin{subfigure}[b]{0.18\textwidth}
        \includegraphics[width=1\textwidth]{images/learn_forget_legend.pdf}
    \end{subfigure}
    }
    \captionsetup[subfigure]{labelformat=parens}
    \begin{subfigure}[b]{0.30\textwidth}
        \includegraphics[width=1\textwidth]
        {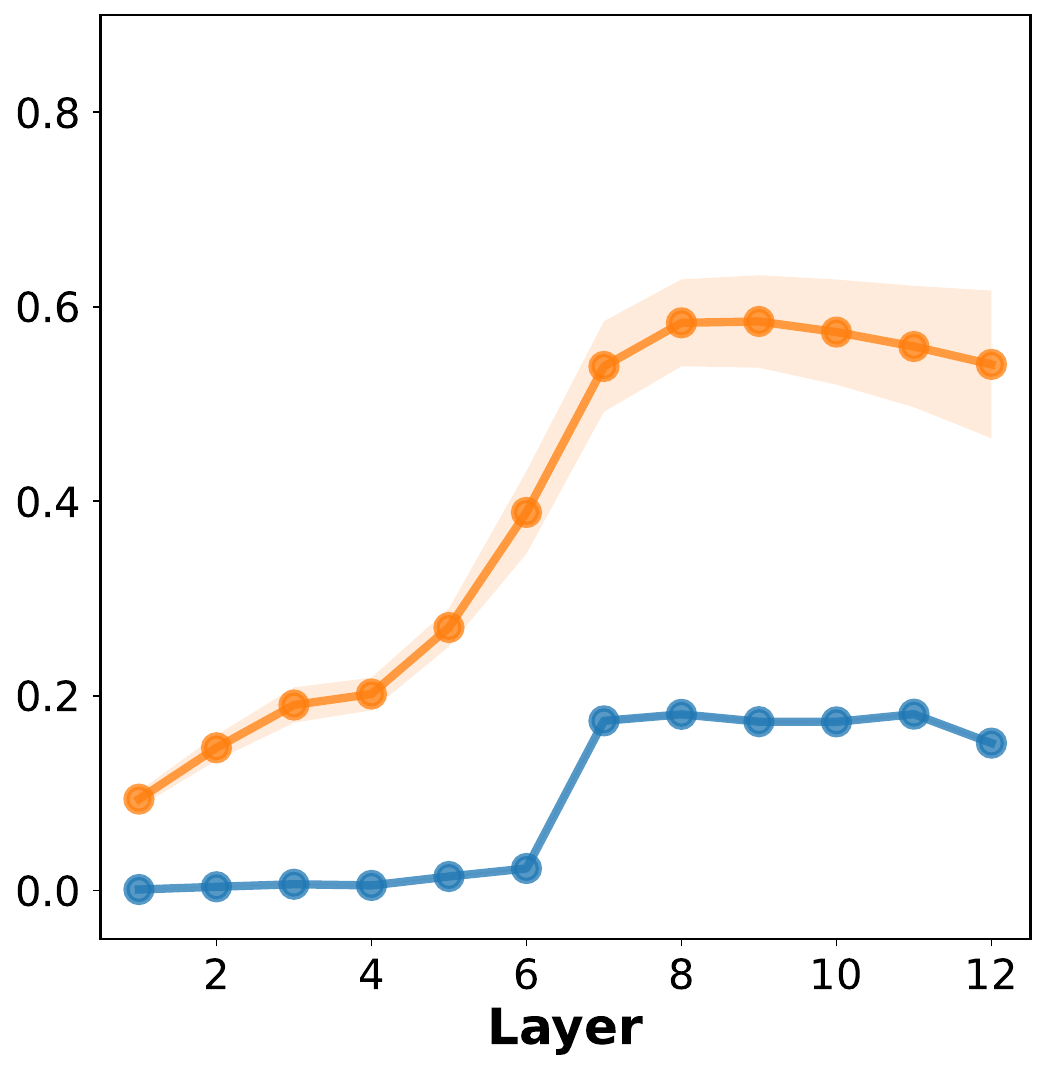}
        \caption{CKA divergence EuroSAT}
        \label{fig:cka_divergence_eurosat}
    \end{subfigure}
    \captionsetup[subfigure]{labelformat=empty}
   \raisebox{7\height}{\begin{subfigure}[b]{0.18\textwidth}
        \includegraphics[height=0.17\textwidth]{images/cka_legend.pdf}
    \end{subfigure}
    }
   \caption{\texttt{learning}, \texttt{forgetting} and \texttt{cka\:divergence} values for ViT model pretrained on ImageNet and finetuned on EuroSAT on reconstruction and classification task.}
\end{figure}

\begin{figure}[ht]
\centering
    \begin{subfigure}[b]{0.30\textwidth}
        
        \includegraphics[width=1\textwidth]{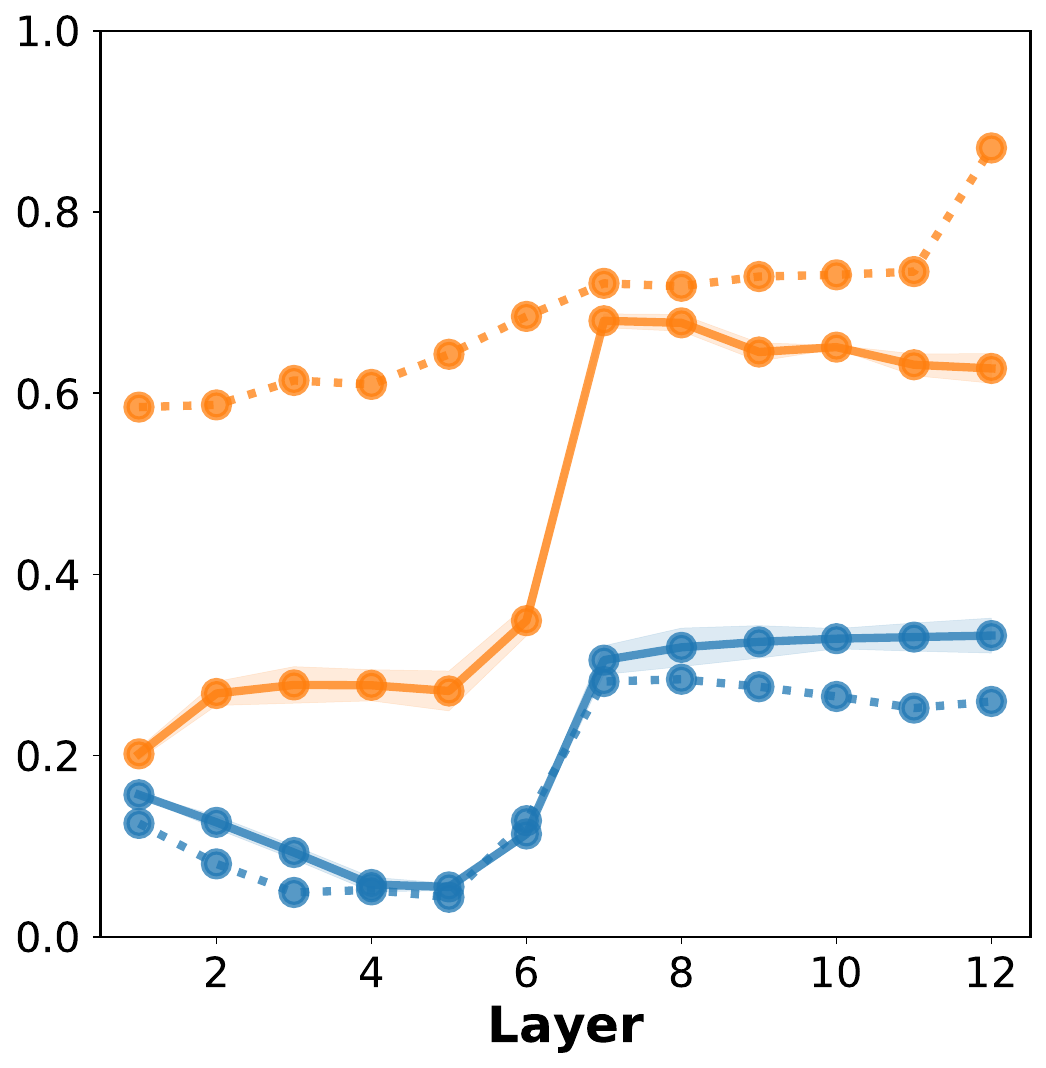}
        \caption{Learning and forgetting Oxford-IIIT Pet}
        \label{fig:learning_forgetting_pets}
    \end{subfigure}
    \captionsetup[subfigure]{labelformat=empty}
   \raisebox{3.5\height}{\begin{subfigure}[b]{0.18\textwidth}
        \includegraphics[width=1\textwidth]{images/learn_forget_legend.pdf}
    \end{subfigure}
    }
    \captionsetup[subfigure]{labelformat=parens}
    \begin{subfigure}[b]{0.30\textwidth}
        \includegraphics[width=1\textwidth]
        {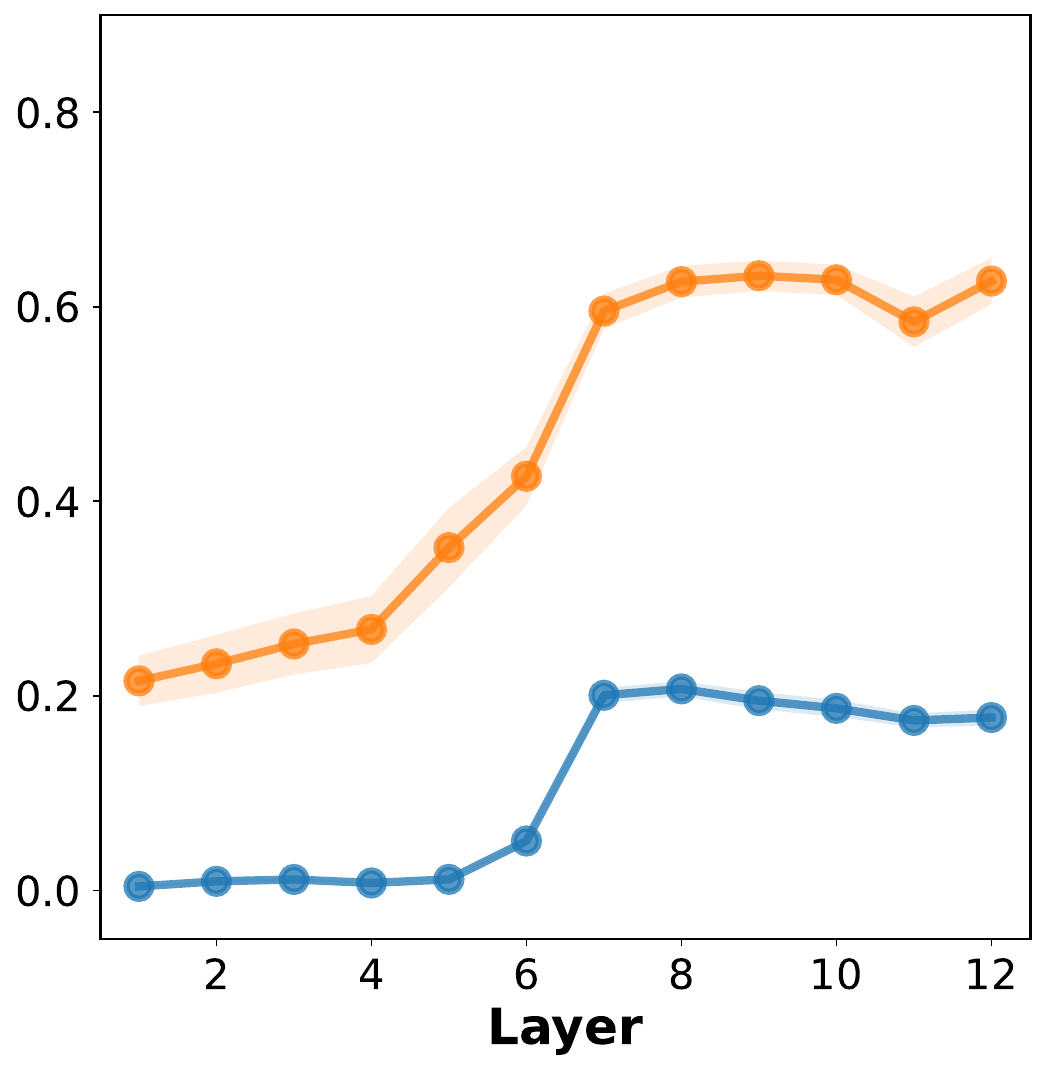}
        \caption{CKA divergence Oxford-IIIT Pet\\\hspace{\textwidth}}
        \label{fig:cka_divergence_pets}
    \end{subfigure}
    \captionsetup[subfigure]{labelformat=empty}
   \raisebox{7\height}{\begin{subfigure}[b]{0.18\textwidth}
        \includegraphics[height=0.17\textwidth]{images/cka_legend.pdf}
    \end{subfigure}
    }
   \caption{\texttt{learning}, \texttt{forgetting} and \texttt{cka\:divergence} values for ViT model pretrained on ImageNet and finetuned on Oxford-IIT Pet dataset on reconstruction and classification task.}
\end{figure}

\newpage

\subsection{Correlation between learning and forgetting}

In Figures \ref{fig:learning_forgetting_cifar10},\ref{fig:learning_forgetting_cifar100},\ref{fig:cifar10_diff_pretrainings},\ref{fig:cifar100_diff_pretrainings} we reported \texttt{learning}, \texttt{forgetting} metrics for different settings. Even if they may show different trends, they are significantly correlated. 
\begin{itemize}
    \item Imagenet $\rightarrow$ CIFAR100 classification: 0.97
    \item Imagenet $\rightarrow$ CIFAR10 classification: 0.97
    \item Imagenet $\rightarrow$ CIFAR100 reconstruction: 0.88
    \item Imagenet $\rightarrow$ CIFAR10 reconstruction: 0.86
    \item CIFAR100 $\rightarrow$ CIFAR10 classification: 0.98
    \item CIFAR10 $\rightarrow$ CIFAR100 classification: 0.99
    \item Random $\rightarrow$ CIFAR10 classification: 0.83
    \item Random $\rightarrow$ CIFAR100 classification: 0.84
\end{itemize}

The correlation however decreases for the reconstruction tasks and for the models trained from scratch. The two measures therefore can be different even if they are correlated, and it is important to take into consideration both of them in future analysis.

\newpage

\section{Invariance Flow Matrix CIFAR100}
\label{app:comp_cifar100}

\begin{figure}[ht]
\centering
    \begin{subfigure}{0.35\textwidth}
        
        \includegraphics[width=\textwidth]{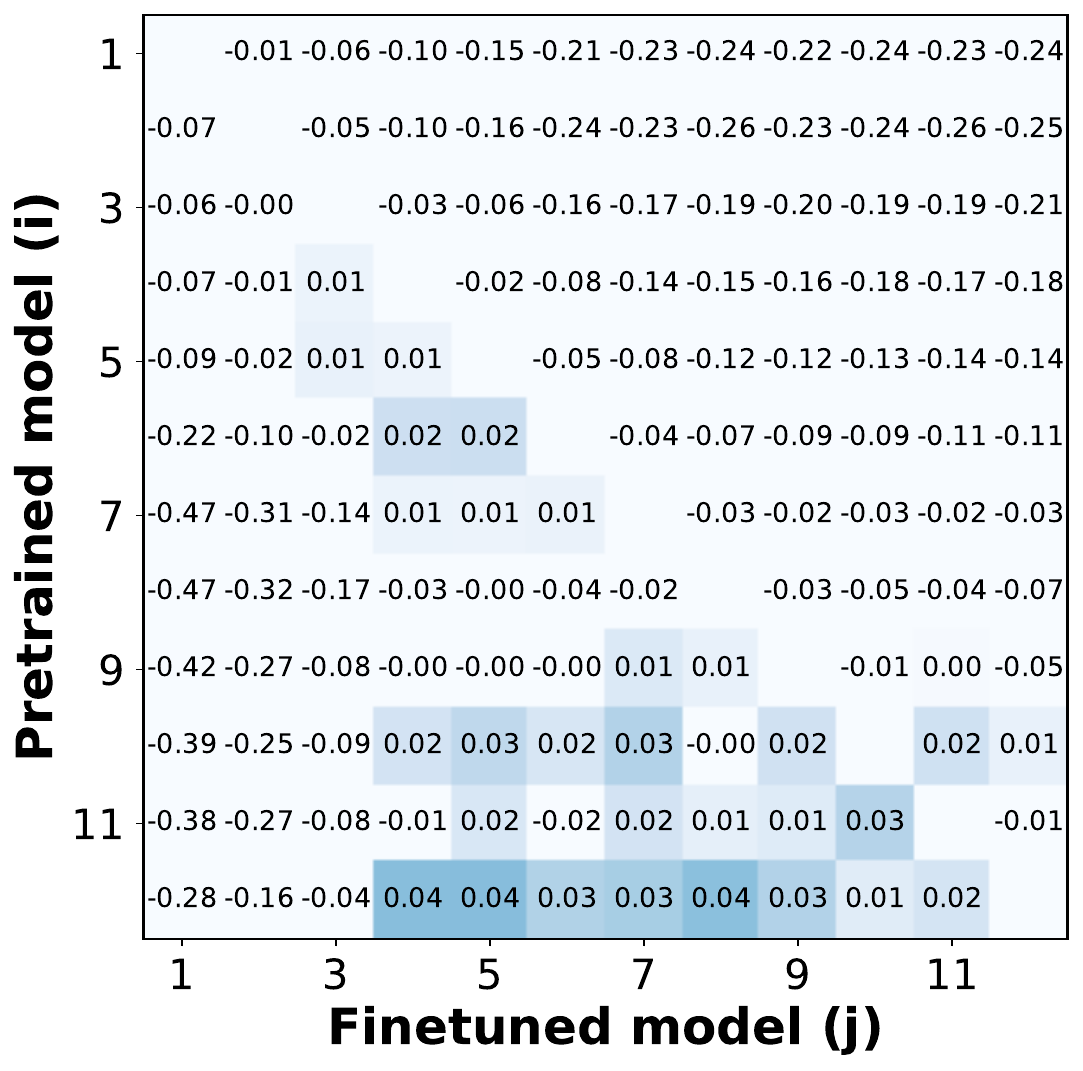}
        \caption{Classification Task}
        \label{fig:comp_class_cifar100}
    \end{subfigure}
    \hspace{0.15\textwidth}
   \begin{subfigure}{0.35\textwidth}
    
        \includegraphics[width=\textwidth]{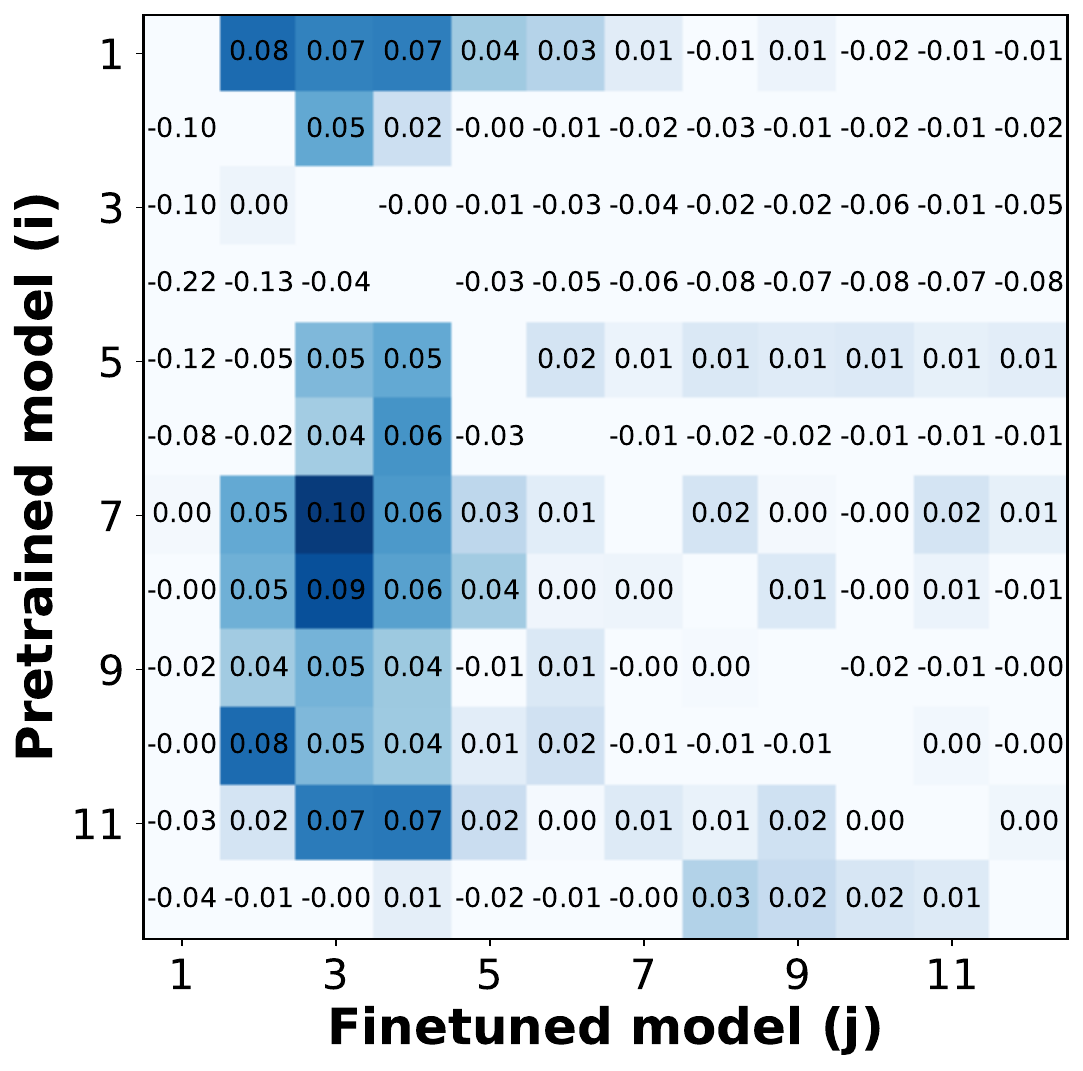}
        \caption{Reconstruction Task}
        \label{fig:comp_recon_cifar100}
    \end{subfigure}
   \caption{Invariance Flow Matrix for finetuned model on classification/reconstruction task of CIFAR100, pretrained on ImageNet.}
\end{figure}

\section{Learning and Forgetting Dynamics additional results}

\begin{figure}[ht]
\centering
    \begin{subfigure}[t]{0.35\textwidth}
        \vskip 0pt
        
        \includegraphics[width=\textwidth]{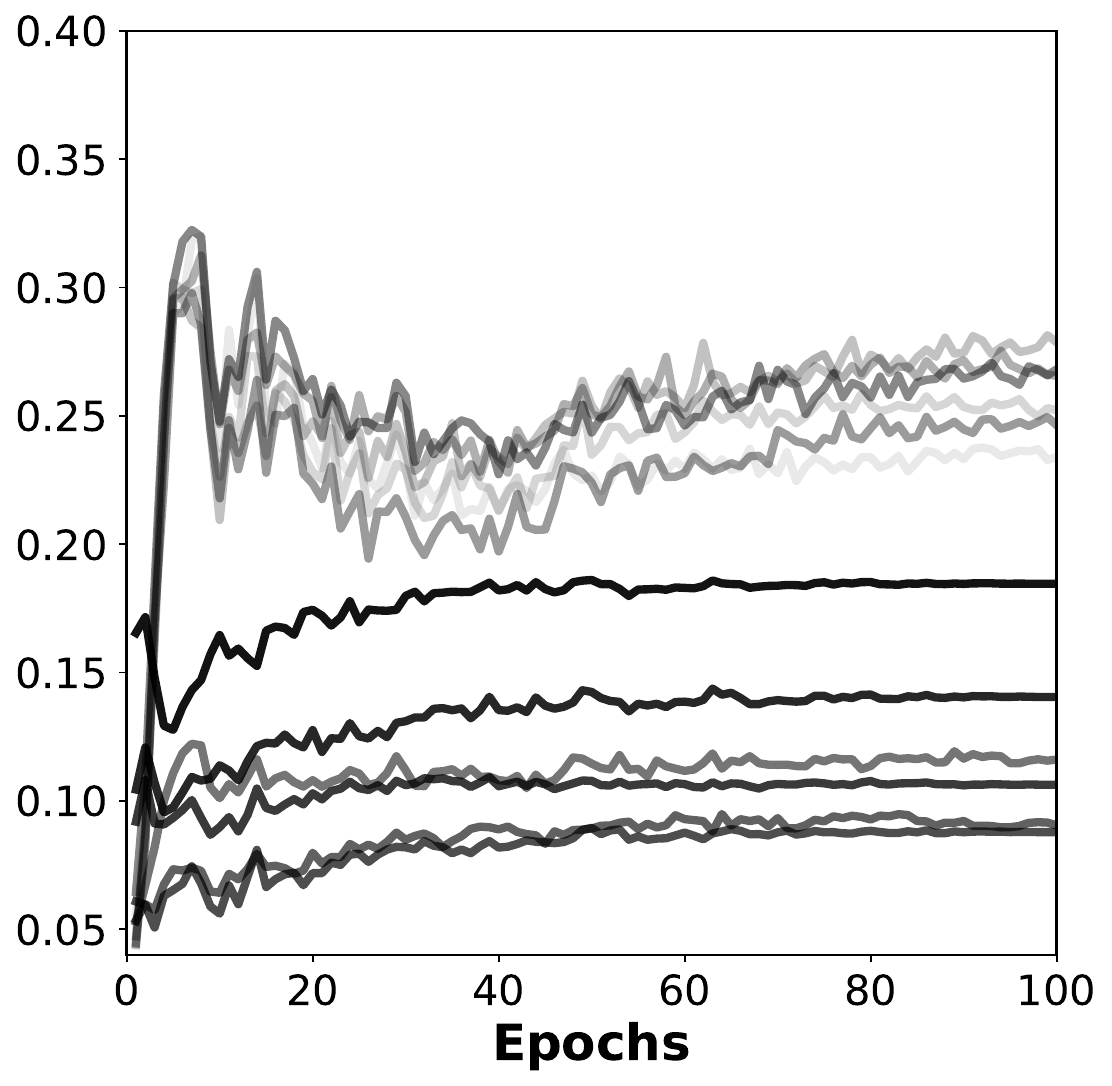}
        \caption{\texttt{learning} during finetuning}
        \label{fig:during_learn_cifar100}
    \end{subfigure}
    \hspace{0.15\textwidth}
   \begin{subfigure}[t]{0.35\textwidth}
        \vskip 0pt
        
        \includegraphics[width=\textwidth]{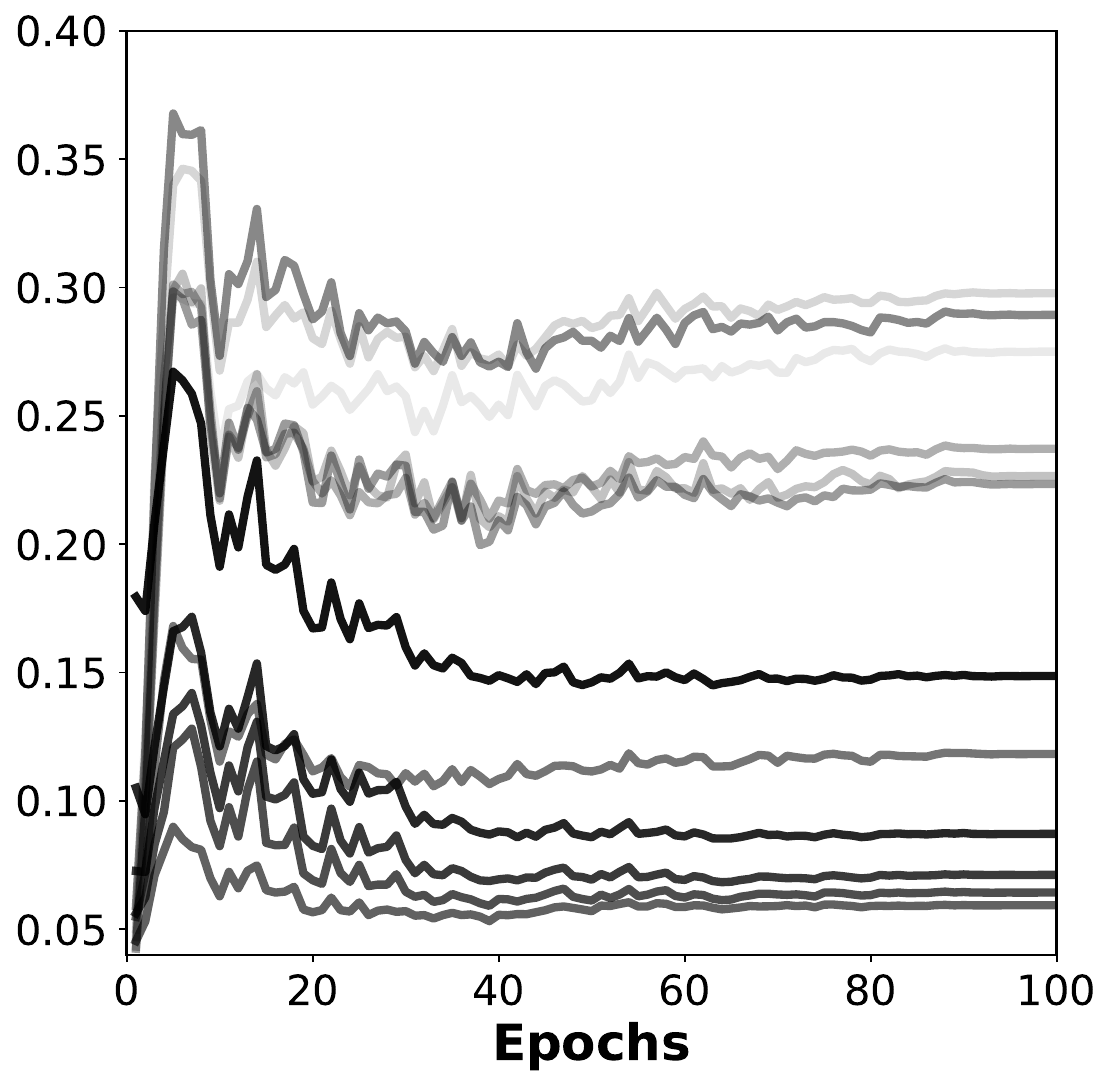}
        \caption{\texttt{forgetting} during finetuning}
        \label{fig:during_forget_cifar100}
    \end{subfigure}
    \captionsetup[subfigure]{labelformat=empty}
    \begin{subfigure}[t]{0.35\textwidth}
        
        \vskip 0pt
        \includegraphics[width=\textwidth]{images/during_legend.pdf}
        
    \end{subfigure}
   \caption{\texttt{learning} and \texttt{forgetting} during finetuning on classification task of CIFAR100 of a model pretrained on ImageNet. We observe a peak of the two metrics in earlier epochs.Only the \texttt{learning} of earlier layer does not exhibit a peak.}
\end{figure}

\begin{figure}[ht]
\centering
    \begin{subfigure}[t]{0.35\textwidth}
        \vskip 0pt
        
        \includegraphics[width=\textwidth]{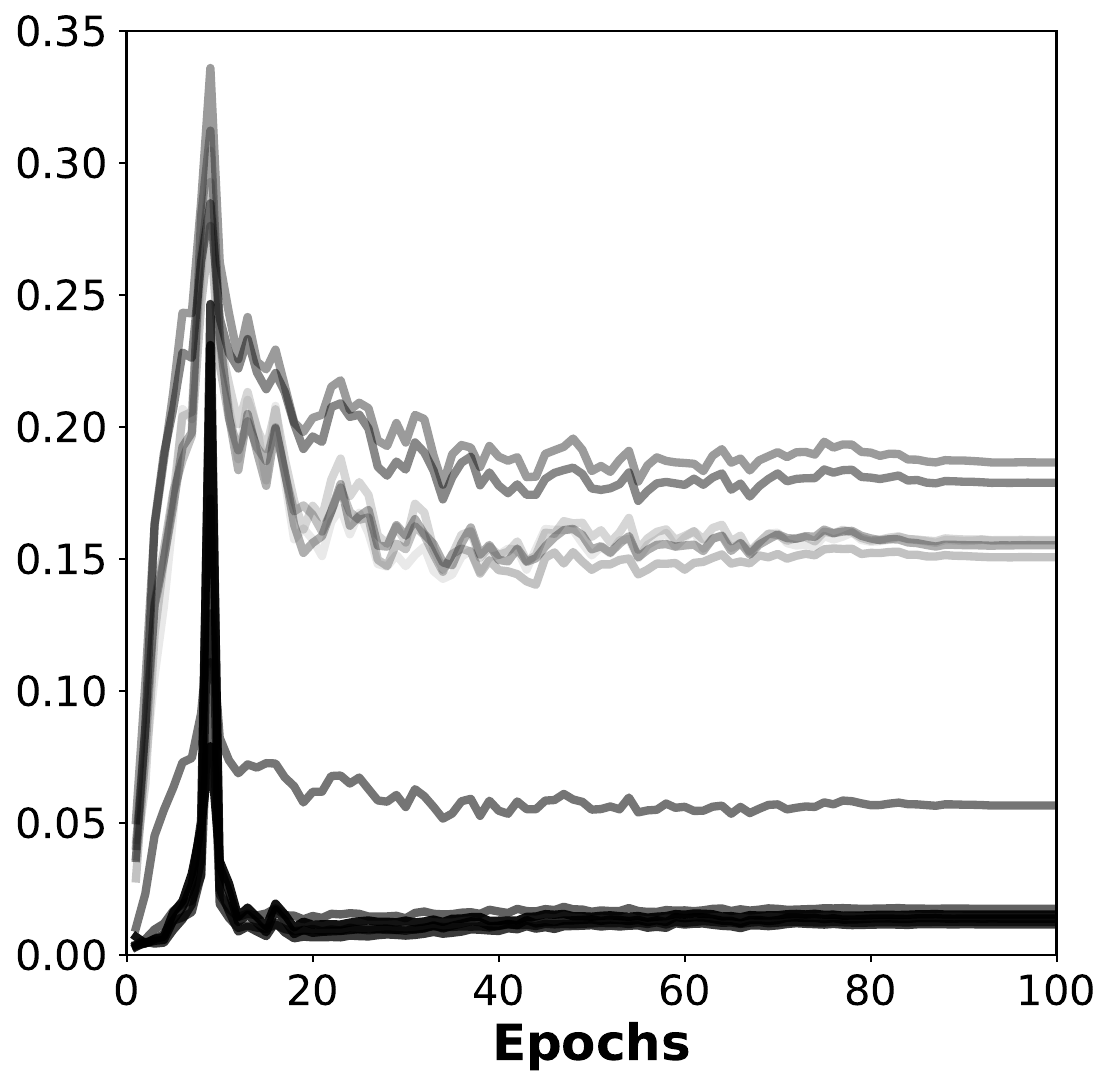}
        \caption{CKA divergence during finetuning CIFAR10}
        \label{fig:during_cka}
    \end{subfigure}
    \hspace{0.15\textwidth}
   \begin{subfigure}[t]{0.35\textwidth}
        \vskip 0pt
        
        \includegraphics[width=\textwidth]{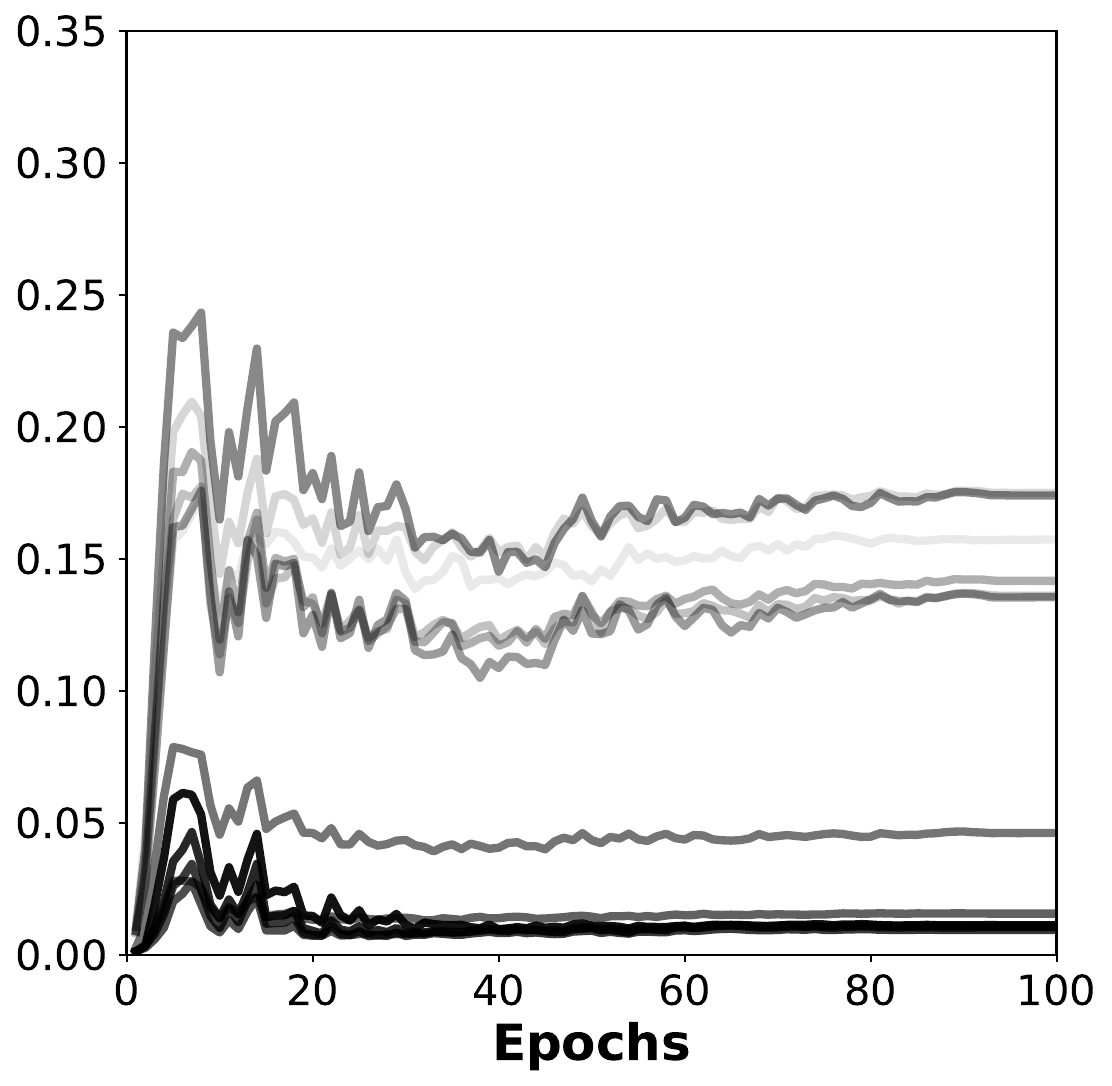}
        \caption{CKA divergence during finetuning CIFAR100}
        \label{fig:during_cka_cifar100}
    \end{subfigure}
    \captionsetup[subfigure]{labelformat=empty}
    \begin{subfigure}[t]{0.35\textwidth}
        
        \vskip 0pt
        \includegraphics[width=\textwidth]{images/during_legend.pdf}
        
    \end{subfigure}
   \caption{\texttt{cka\:divergence} during finetuning on classification task of CIFAR10 and CIFAR100 of a model pretrained on ImageNet. We observe a peak in earlier epochs.}
\end{figure}

\begin{figure}[ht]

  \centering
  \includegraphics[width=0.5\columnwidth]{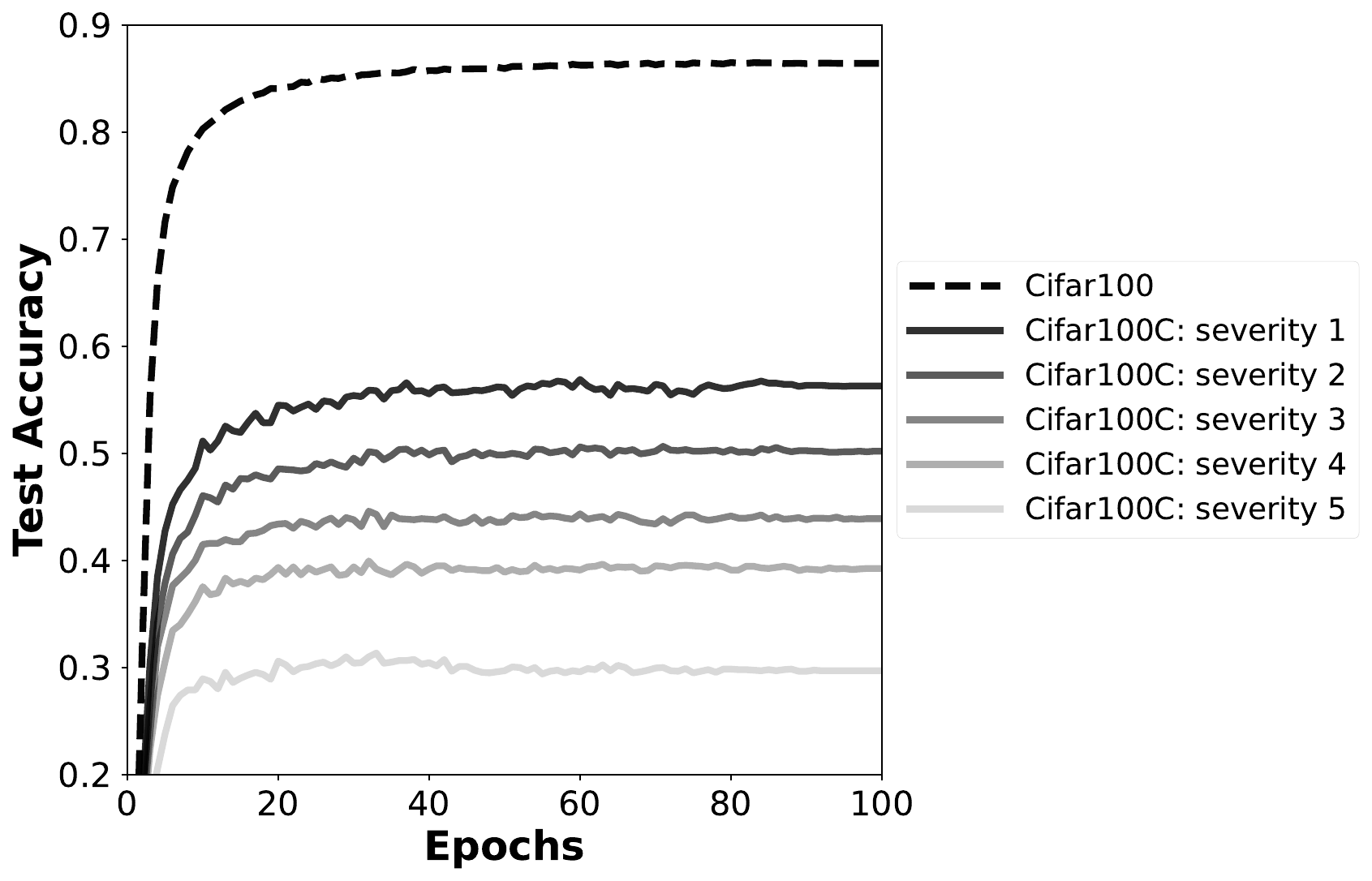}
  \caption{Test accuracy during finetuning on classification task of CIFAR100 of a model pretrained on ImageNet. Different lines correspond to different corruption level of the original CIFAR10 test set. While the accuracy on the standard test set increase, the accuracy on corrupted dataset does not always increase, in particular for dataset with high level of corruption(\ie\ severity 3, 4, 5). In these cases the accuracy has a peak on earlier epochs.}
  \label{fig:during_acc_cifar100}

\end{figure}

\newpage
\subsection{Correlation with robustness}
\label{app:corr_robust}
For this experiment we explored 352 hypotesis using aggregate metrics. We varied:
\begin{itemize}
    \item Layer considered: only one layer, first $n$ layers or last $n$ layers.
    \item \texttt{learning} and \texttt{forgetting} operation: addition, subtraction, only \texttt{learning} or only \texttt{forgetting}
    \item Aggregate layers operation: mean, standard deviation, minimum or maximum.
\end{itemize}

Similarly to \texttt{learning} or only \texttt{forgetting} for the \texttt{cka\:divergence} we vearied the layers considered and the aggregate layers operations for a total number of 88 combinations
To test the accuracy of the model we use the standard test set of CIFAR10 and CIFAR100, and we use 1000 randomly sampled inputs for each level of corruption.

\begin{figure}[ht]

  \centering
  \includegraphics[width=0.5\columnwidth]{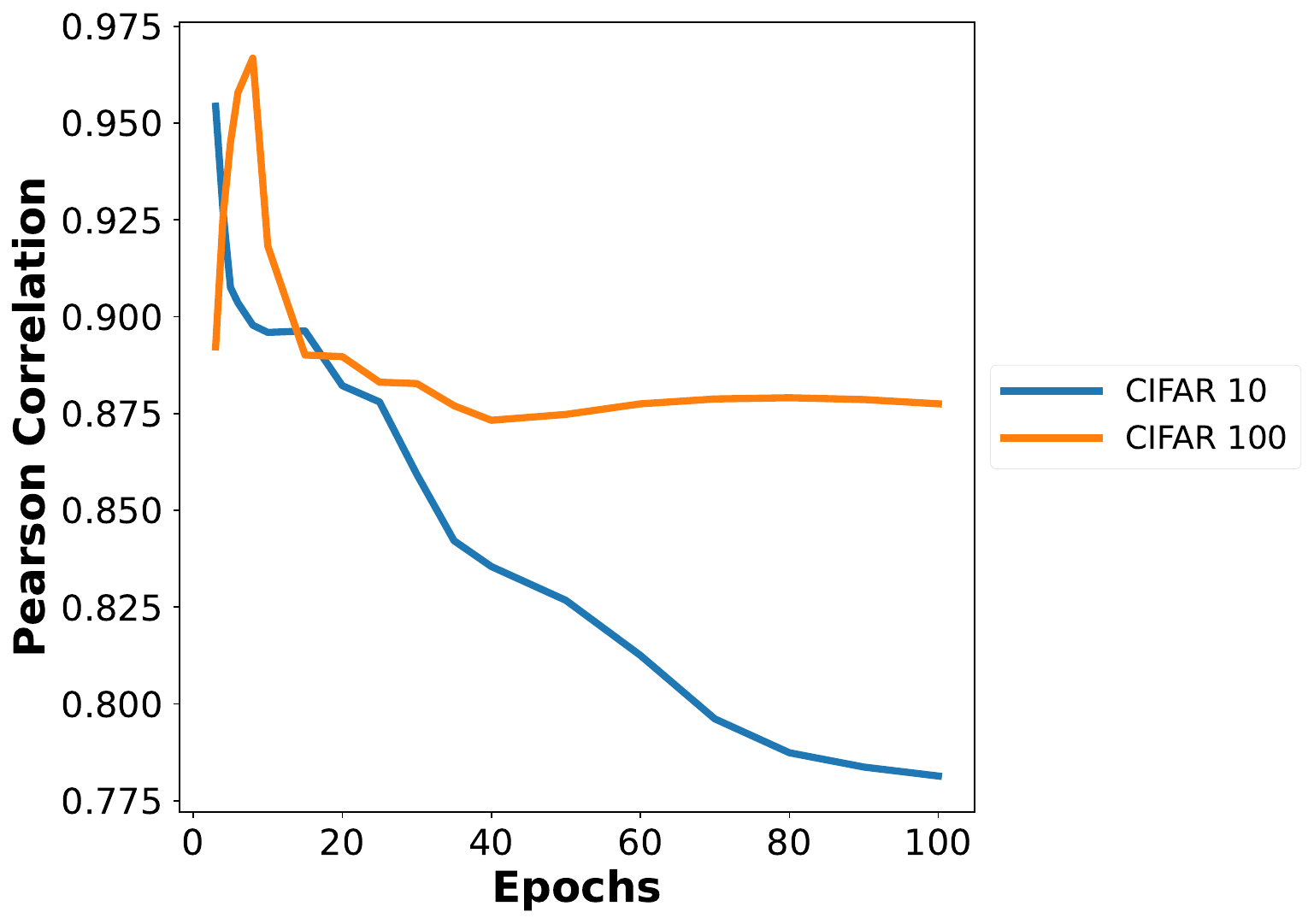}
  \caption{Correlation between average accuracy on CIFAR10C/CIFAR100C datasets and standard deviation of \texttt{forgetting} across layers 2-12 during training. Each point on the graph represents the correlation between the average of the accuracies on the CIFAR10C/CIFAR100C datasets and the standard deviation of \texttt{forgetting} across layers 2-12 considered up to the epoch indicated on the x-axis. For example at epoch $n$ the correlation is computed between $[avg\_acc\_epoch\_0, avg\_acc\_epoch\_1,..., avg\_acc\_epoch\_n]$ and $[std\_forgetting\_epoch\_0, std\_forgetting\_epoch\_1, ..., std\_forgetting\_epoch\_n]$.}
  \label{fig:corr_analysis}

\end{figure}

\begin{figure}[ht]

  \centering
  \includegraphics[width=0.5\columnwidth]{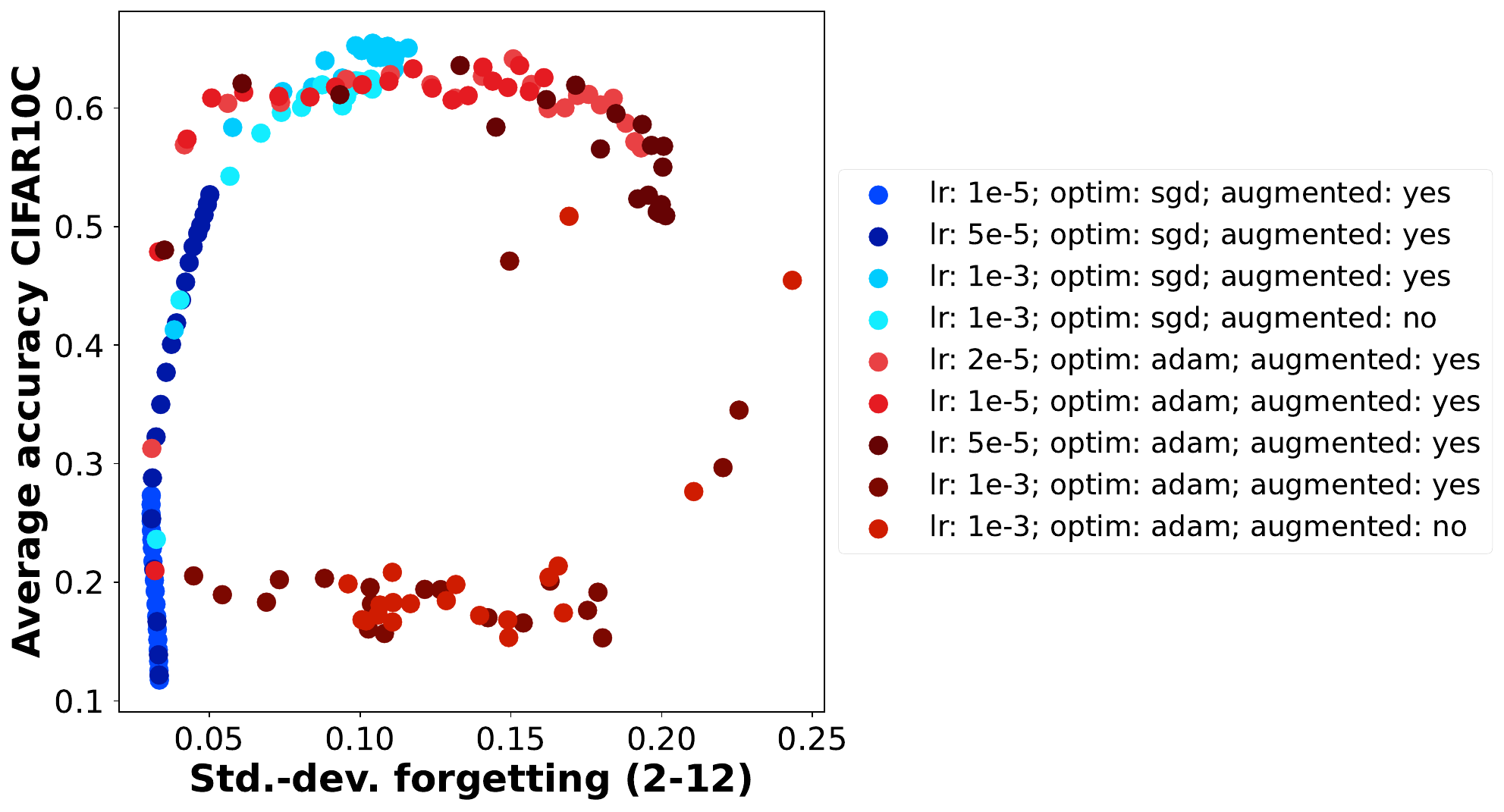}
  \caption{Average accuracy on Cifar10C datasets and standard deviation of \texttt{forgetting} across layers 2-12 of 20 epochs of different models while training on classification of CIFAR10 with $sgd$ and $adam$ optimizers.}
  \label{fig:acc_forg}

\end{figure}

\end{document}

%% file: math_commands.tex
\usepackage{amsmath,amsfonts,bm}

\def\eqref#1{equation~\ref{#1}}

\def\1{\bm{1}}

\DeclareMathAlphabet{\mathsfit}{\encodingdefault}{\sfdefault}{m}{sl}
\SetMathAlphabet{\mathsfit}{bold}{\encodingdefault}{\sfdefault}{bx}{n}

